\documentclass[12pt,onecolumn,draftclsnofoot]{IEEEtran}

\usepackage[encapsulated]{CJK}
\usepackage{ucs}
\usepackage[utf8x]{inputenc}
\usepackage[cmex10]{amsmath}
\usepackage{amsmath,amssymb,amscd,bbm,amsthm,mathrsfs,dsfont}
\usepackage{algorithmic,algorithm}
\usepackage{mdwmath}
\usepackage{mdwtab}
\usepackage{bm,upgreek}
\usepackage{cite}
\usepackage{rotating,graphics,psfrag,epsfig}
\usepackage{array}
\usepackage{booktabs}
\usepackage{indentfirst}
\usepackage{subfigure}
\usepackage{lipsum,fancyhdr,lastpage,refcount}
\usepackage{url}
\usepackage{fixltx2e,dblfloatfix}
\usepackage{blindtext}
\usepackage[T1]{fontenc}
\usepackage{color}
\usepackage[normalem]{ulem}

\IEEEoverridecommandlockouts

\let\oldnl\nl
\newcommand{\nonl}{\renewcommand{\nl}{\let\nl\oldnl}}

\begin{document}

\title{Age of Semantics in Cooperative Communications: To Expedite Simulation Towards Real via Offline Reinforcement Learning}

\author{\IEEEauthorblockN{Xianfu Chen, \emph{Member}, \emph{IEEE}, Zhifeng Zhao, \emph{Member}, \emph{IEEE}, Shiwen Mao, \emph{Fellow}, \emph{IEEE}, Celimuge Wu, \emph{Senior Member}, \emph{IEEE}, Honggang Zhang, \emph{Senior Member}, \emph{IEEE}, and Mehdi Bennis, \emph{Fellow}, \emph{IEEE}}


\thanks{X. Chen is with the VTT Technical Research Centre of Finland, Oulu, Finland (email: xianfu.chen@vtt.fi).}

\thanks{Z. Zhao is with the Zhejiang Lab, and also with the College of Information Science and Electronic Engineering (ISEE), Zhejiang University, Hangzhou, China (email: zhaozf@zhejianglab.com).}

\thanks{S. Mao is with the Department of Electrical and Computer Engineering, Auburn University, Auburn, AL, USA (email: smao@ieee.org).}

\thanks{C. Wu is with the Graduate School of Informatics and Engineering, University of Electro-Communications, Tokyo, Japan (email: celimuge@uec.ac.jp).}

\thanks{H. Zhang is with the Zhejiang Lab, and also with the College of Information Science and Electronic Engineering (ISEE), Zhejiang University, Hangzhou, China (email: honggangzhang@zju.edu.cn).}

\thanks{M. Bennis is with the Centre for Wireless Communications, University of Oulu, Finland (email: mehdi.bennis@oulu.fi).}

\thanks{This work has been submitted to the IEEE for possible publication. Copyright may be transferred without notice, after which
this version may no longer be accessible.}

}

\maketitle

\begin{abstract}

The age of information metric fails to correctly describe the intrinsic semantics of a status update.
In an intelligent reflecting surface-aided cooperative relay communication system, we propose the age of semantics (AoS) for measuring semantics freshness of the status updates.
Specifically, we focus on the status updating from a source node (SN) to the destination, which is formulated as a Markov decision process (MDP).
The objective of the SN is to maximize the expected satisfaction of AoS and energy consumption under the maximum transmit power constraint.
To seek the optimal control policy, we first derive an online deep actor-critic (DAC) learning scheme under the on-policy temporal difference learning framework.
However, implementing the online DAC in practice poses the key challenge in infinitely repeated interactions between the SN and the system, which can be dangerous particularly during the exploration.
We then put forward a novel offline DAC scheme, which estimates the optimal control policy from a previously collected dataset without any further interactions with the system.
Numerical experiments verify the theoretical results and show that our offline DAC scheme significantly outperforms the online DAC scheme and the most representative baselines in terms of mean utility, demonstrating strong robustness to dataset quality.

\end{abstract}


\section{Introduction}
\label{intr}

Cooperative relay communications have exhibited high potentials in expanding the system coverage and capacity \cite{Wang13}.
Recently, hybrid relay systems are emerging to further enhance the relaying performance, where intelligent reflecting surfaces (IRSs) are deployed to improve the propagation conditions \cite{ZKang22}.
Specifically, an IRS consists of a large number of passive reflecting elements that adapt the propagation environment by tuning the amplitudes and/or phase-shifts \cite{Bjor20}.
Without the need of any radio-frequency chains, an IRS is able to achieve cost and energy-efficient communications.
In this paper, we study an IRS-aided cooperative relay communication system, where a source node (SN) updates to the destination through sampling the status of an underlying process.
Typical scenarios include the real-time monitoring in complex smart manufacturing \cite{Noor22} and the video analytics in autonomous driving \cite{Wang21}, to mention a few, where the fresh information and semantics of the process of interest is crucial from the perspective of the destination.
Let us take the vehicle detection and tracking as an illustrative example, where the intelligent camera (IC; i.e., the SN) responds to the remote control unit (RCU; i.e., the destination) \cite{Du20}.
Each image captured by the IC can be considered to be composed of the target vehicle part and the background part.
In order to save communication resource, the IC employs a semantic extraction module to compress the image, and the compressed data is sent to the RCU \cite{Kang22}.
Afterwards, the RCU performs semantic reconstruction to recover the image for inference, the result from which can be used as the input to high-level applications (e.g., traffic accident surveillance).
With the inference result from the RCU and the situational awareness, the IC decides whether or not to, for example, pan, tilt or zoom in/out to capture a new image in a relevant region.

\subsection{Related Works and Motivation}
\label{lite}

Maintaining fresh information of the process of interest at the destination requires the SN to send the time-stamped status updates, which motivates the introduction of age of information (AoI) \cite{Kaul12, Yates21}.
At the destination, AoI quantifies the time lag since the generation of the most recently received update.
In the literature, most efforts have been focused on exploring AoI in single-hop communication systems \cite[and the references therein]{Kaul12, ChenX20, ChenX22, Qian20, Abd20, Ahani22, Yates21}.
It remains daunting to minimize AoI in cooperative relay communication systems, where the path selection and the resource constraint noticeably expand the dimensionality.
In \cite{Talak17}, Talak et al. studied simple stationary policies to minimize AoI for multi-hop networks under general interference constraints.
In \cite{Farazi18}, Farazi et al. derived the lower bounds on peak and average AoI in multi-hop wireless networks with explicit channel contention.
In \cite{He22}, He et al. proposed to minimize the maximum average AoI in a multi-hop Internet-of-Things network, which was formulated as a mixed integer linear programming problem jointly optimizing beamforming vector and routing.
In \cite{Lou22}, Lou et al. developed a linearized approximate algorithm and a polynomial time algorithm to optimize AoI in a multi-hop wireless network.
In \cite{Liu22}, Liu et al. proved that the peak/average AoI minimization in multi-path communications turned out to be roughly equivalent to minimizing the maximum delay, which was leveraged to design a general approximation solution.
However, the neglect of the tight coupling between decision-makings and uncertainties leads to suboptimal AoI in dynamic relay communication systems.

In a cooperative relay communication system, the uncertainties originate from not only the variations during status update transmissions (e.g., the temporally changing channel gain and system topology), but also the randomness in resource availabilities (e.g., the transmit power and computation capability) \cite{Chen19J}.
Markov decision process (MDP) provides a mathematical framework for controlling actions (i.e., decision-makings) under uncertainties over the time horizon \cite{Rich98}.
Using a constrained MDP, Gu et al. investigated the problem of age minimization for a two-hop relay system under a resource budget \cite{Gu21}.
In \cite{Tripathi21}, Tripathi et al. converted the AoI optimization into network stability problems, for which the Lyapunov drift was used to find the scheduling and routing policies.
The Lyapunov technique does not rely on the MDP statistics but only constructs an approximately optimal policy.
Reinforcement learning (RL) has been successful in solving an MDP without a priori statistical information \cite{ChenX20, ChenX22, Abd20}.
To our best knowledge, there is yet no comprehensive attempt to unleash the power of RL for AoI optimization in cooperative relay communication systems.
As we shall illustrate, this work will go even further and concentrate on the following two aspects.
\begin{enumerate}
  \item \emph{AoI versus Semantics Freshness:} We note that the majority of the prior research has been directed at optimizing the information freshness.
      There is a huge potential for the SN to boost the resource efficiency of status update transmissions if the intrinsic semantics of the process of interest can be extracted \cite{Du20, Kang22}.
      In a cooperative relay communication system, the ultimate goal of the destination is to promptly grasp the inference (e.g., the inference result from each recovered image for traffic accident surveillance as in the illustrative example above) from the status updates.
      In line with the definition, even when the destination has the perfect inference of the current process status, AoI keeps increasing until a new update is received from the SN.
      In \cite{Sun20}, Sun et al. validated that the expected estimation performance from the mean squared error (MSE)-minimum sampling is much better than the AoI-optimal sampling for the Wiener process with random delay.
      To address such a shortcoming of AoI, Maatouk et al. introduced the age of incorrect information (AoII), which incorporated the content of the status updates into the design of a transmission policy within the MDP framework \cite{Maat20}.
      However, we still lack a metric that reveals the relationship between the semantics of the process of interest at the SN and the timely inference from a status update at the destination.

  \item \emph{Online versus Offline RL:} In a simulated system, an RL agent utilizes the newly collected interaction experiences to update the control policy parameters.
      Meanwhile, the experiences come from implementing the control policy to be optimized.
      The repeated alternate between updating the control policy parameters and collecting the interaction experiences over the time horizon is considered as the online RL training.
      This falls into the ``chicken and egg'' paradox, which restricts the application of online RL to a real system.
      On one hand, the continuous online collection of interaction experiences is extremely challenging.
      On the other hand, the random actions from the RL agent during the trial-and-error exploration are dangerous for operating a system \cite{Xu21}.
      Accordingly, in a real cooperative relay communication system, the control policy of the SN has to be pre-trained offline.
      The offline RL training leverages a static dataset, which is composed of a number of interaction experiences (e.g., from the historical system operations) \cite{Levine20}.
      We refer to each interaction experience as a tuple of current system state, action, immediate utility and subsequent system state.
      Towards this direction, one theme of work was centered on learning the control policy by constraining the feasible action space to the support of the dataset \cite{Xu21, Fujimoto19, Kumar19, Siegel20}.
      Another recent theme is aimed at preventing the extrapolation errors attributed to the out-of-distribution (OOD) actions, which are those that do not appear under the same current system state of an interaction experience in the dataset \cite{Wang20, Kumar20, Xu22}.
      However, the state-of-the-art results on offline RL are either not applicable to discrete action settings or are highly sensitive to the dataset quality.
\end{enumerate}

\subsection{Contribution and Structure}

In this paper, we shall address the above challenges by designing an offline RL scheme for the optimization of semantics freshness in an IRS-aided cooperative relay communication system.
The SN updates the process of interest to the destination over the infinite time horizon through sampling the status.
More specifically, the SN compresses the status updates using semantic extraction to derive the semantic samples, which are sent to the destination with the help of multiple relay stations (RSs) and an IRS.
Accounting for the source constraints, the SN has to learn to choose the sampling and the RS selection actions with the perception of system uncertainties.
We summarize the unique technical contributions from this work as follows.
\begin{itemize}
  \item We define a novel metric, which is termed as age of semantics (AoS), to connect the semantics of the process of interest at the SN and the timely inference at the destination.
      In particular, the process status is modelled using a discrete Markov chain, while the inference from status update reconstruction at the destination follows a stochastic process.
      Different from AoI and AoII, AoS extends the information freshness to the semantics freshness, which is calculated as the time duration since the perfect inference of the current process status.
  \item We formulate the problem of semantics freshness optimization as a discrete MDP, the objective of which is to maximize the expected discounted utility.
      Without the requirement of system uncertainty statistics, we first propose an online deep actor-critic (DAC) scheme, which applies the on-policy temporal difference (TD) method \cite{Rich98}.
      From online interactions with the system, the DAC scheme enables the SN to learn to approach the optimal control policy, which maps each system state to a distribution of the sampling and RS selection actions.
  \item For an IRS-aided cooperative relay cooperative system in practise, the continuous online interactions are impossible for the SN, which motivates us to further propose an offline data-driven DAC scheme.
      Without any interactions with the system, the SN trains the offline DAC scheme by making use of only a previously collected static dataset of interaction experiences.
      Numerical experiments demonstrate the robustness of our proposed offline DAC scheme to the dataset quality by exceeding the online advantage actor critic (A2C) scheme \cite{Mnih16} with a small margin, and significantly outperforming the most representative existing offline deep RL scheme, namely, conservative Q-learning (CQL) \cite{Kumar20}.
\end{itemize}

The rest of this paper is structured as follows.
In the next section, we introduce the system model and elaborate the AoS metric.
In Section \ref{prob}, we formulate the problem of semantics freshness optimization in an IRS-aided cooperative relay system as a discrete MDP and develop an online DAC scheme to approach the optimal control policy.
We also analyze the challenges faced by the online DAC scheme.
In Section \ref{offline_scheme}, we propose an offline DAC scheme, which enables the SN to learn the control policy from a static dataset of interaction experiences.
In Section \ref{simu}, we present the numerical experiments and discuss the evaluation results.
Finally, we draw the conclusions in Section \ref{conc}.

\section{Models and Assumptions}
\label{sysm}

\begin{figure}[t]
  \centering
  \includegraphics[width=16pc]{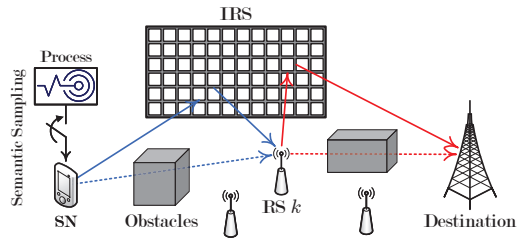}
  \caption{Illustration of an IRS-aided cooperative relay communication, in which the SN updates the process of interest to the destination through sampling the status over the infinite time horizon.}
  \label{systMode}
\end{figure}

As illustrated in Fig. \ref{systMode}, we study an IRS-aided cooperative relay communication system, where the SN updates the process of interest to the destination through transmitting the semantic samples over the discrete time slots.
The SN cannot reach the destination directly due to the limited coverage and a set $\mathcal{K} = \{1, \cdots, K\}$ of RSs are deployed to extend the communication range of SN.
The RSs are half-duplex relays in a decode-forward mode.
All of the SN, the RSs and the destination have a single antenna.
In the system, an IRS with $I$ reflecting elements enhances the transmission links from SN to RSs as well as the links from RSs to destination, which can be possibly blocked by the dynamic obstacles.
Each IRS element has a smaller size than the wavelength and hence scatters the incoming signal equally in all directions \cite{Bjor20}.

\begin{figure}[t]
  \centering
  \includegraphics[width=16.9pc]{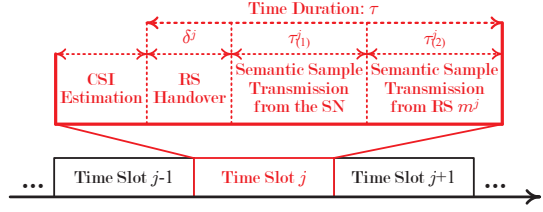}
  \caption{Time slot structure.}
  \label{timeSlot}
\end{figure}

We index each time slot by an integer $j \in \mathds{N}_+$ and all time slots are assumed to be of equal duration.
To facilitate the semantic sample transmissions, a time slot is further divided into four sub-slots as in Fig. \ref{timeSlot}.
\begin{enumerate}
  \item \emph{Fixed channel estimation sub-slot} is used to estimate the channel state information (CSI) of all transmission links.
  \item \emph{RS handover sub-slot} is triggered for the SN once the selected RS is different from the previously associated one \cite{Cho09}.
  \item \emph{First flexible transmission sub-slot} is used by the SN to transmit a semantic sample to the selected RS, which is aided by the IRS.
  \item \emph{Second dynamic transmission sub-slot} is occupied by the selected RS to decode and forward the received semantic sample to the destination with the aid of the IRS.
\end{enumerate}
At the end of the time slot, the destination reports the inference result back to the SN when the status update is reconstructed from the received semantic sample.
We assume that the inference result is reported using the reverse link via the associated RS, the time consumption of which is negligible because of the small data size \cite{Wang21, Du20}.
For convenience, we designate $\tau$ as the constant sum of the time durations of handover, first flexible transmission and second dynamic transmission sub-slots during one time slot.

\subsection{IRS-Aided Relaying}

Let $n^j \in \{0, 1\}$ define the sampling action of the SN at each time slot $j$, where $n^j = 1$ if the SN decides to sample the process status $X^j$ and otherwise, $n^j = 0$.
When $n^j = 1$, a fresh status update is generated and compressed with the semantic extraction, which outputs a semantic sample of size $\upsilon$.
The energy consumed by status sampling and semantic extraction is assumed to be $\varrho$ \cite{Wang22}.
Following that, the semantic sample is sent to the destination via at most one selected RS $m^j \in \mathcal{K}$.
We denote $m^j \in \widetilde{\mathcal{K}} = \mathcal{K} \cup \{0\}$ as the RS selection action for the SN at a time slot $j$, where in particular, we let $m^j = 0$ when $n^j = 0$ for notational consistency.
Suppose that the status of the process of interest follows a discrete Markov chain, namely, $X^j \in \mathcal{X}$, where $\mathcal{X}$ is a finite state space.
To proceed the transmission of a semantic sample, the SN-RS association has to be established.
Let $y^j \in \mathcal{K}$ denote the SN-RS association state of the SN at time slot $j$, we have $y^j = m^j$ if $m^j \in \mathcal{K}$.
Otherwise, if $n^j = 0$, the SN-RS association state remains as $y^j = y^{j - 1}$.

In this paper, we consider that the RSs transmit with fixed power, while the SN adapts transmit power to the channel conditions \cite{Zlat19}.
We first concentrate on the second dynamic transmission sub-slot during each time slot $j$.
Let $g_{k, (\mathrm{D})}^j \in \mathds{C}$ denote the channel from an RS $k \in \mathcal{K}$ to the destination, while the channels between RS $k$ and the IRS as well as between the IRS and the destination are, respectively, denoted by $\mathbf{g}_{k, (\mathrm{I})}^j \in \mathds{C}^I$ and $\mathbf{g}_{(\mathrm{I}, \mathrm{D})}^j \in \mathds{C}^I$.
Following \cite{Bjor20}, we express the achievable data rate for the IRS-aided uplink from RS $k$ to the destination at slot $j$ as
\begin{align}
     R_{k, (\mathrm{D})}^j = w \cdot
     \log_2\!\!\left(1 + \frac{P_k \cdot \left(\left|g_{k, (\mathrm{D})}^j\right| +
       \zeta \cdot \sum\limits_{i = 1}^I \left|\left[\mathbf{g}_{k, (\mathrm{I})}^j\right]_i \cdot
                                         \left[\mathbf{g}_{(\mathrm{I}, \mathrm{D})}^j\right]_i\right|\right)^2}
                                {w \cdot \sigma^2}\right),
\end{align}
where $w$ is the system frequency bandwidth, $P_k$ is the transmit power of RS $k$, $\zeta \in (0, 1]$ is the fixed amplitude reflection coefficient of the IRS, $\sigma^2$ is the additive noise power spectral density, and $[\cdot]_i$ denotes the $i$-th component of a vector.
If $n^j = 1$ and $m^j \in \mathcal{K}$, the time consumed by transmitting the semantic sample from the selected RS $m^j$ to the destination can be hence calculated as
\begin{align}
   \tau^j_{(2)} = \frac{\upsilon}{R_{m^j, (\mathrm{D})}^j}.
\end{align}
For the first flexible transmission sub-slot during time slot $j$, we then derive the amount of time used to transmit the semantic sample from the SN to the selected RS $m^j$ as
\begin{align}
    \tau^j_{(1)} = \tau - \tau^j_{(2)} - \delta^j,
\end{align}
where $\delta^j = \delta \cdot \mathds{1}_{\{y^j \neq y^{j - 1}\}}$ with $\delta$ being the delay during the occurrence of a handover and $\mathds{1}_{\{\cdot\}}$ denoting an indicator function.
Accordingly, the required transmit power by the SN can be deduced as
\begin{align}
   p^j =
   \frac{w \cdot \sigma^2}
        {\left(\left|g_{m^j}^j\right| + \zeta \cdot \sum\limits_{i = 1}^I \left|\left[\mathbf{g}_{(\mathrm{I})}^j\right]_i \cdot
         \left[\mathbf{g}_{(\mathrm{I}), m^j}^j\right]_i\right|\right)^2}
   \left(2^{\frac{\upsilon}{\tau^j_{(1)} \cdot w}} - 1\right),
\end{align}
where $\mathbf{g}_{(\mathrm{I}), m^j}^j \in \mathds{C}^I$ is the channel from the IRS to RS $m^j$, while $g_{m^j}^j \in \mathds{C}$ and $\mathbf{g}_{(\mathrm{I})}^j \in \mathds{C}^I$ are the channels from the SN to RS $m^j$ and the IRS, respectively.
We denote by $P$ the maximum transmit power of the SN, which constrains that $\forall j$, $p^j \leq P$.

%
%

\subsection{AoS Evolution}

After receiving the semantic sample from the SN at the end of a time slot $j$, the destination reconstructs the status update, which leads to an inference $\widehat{X}^{j + 1}$ of the process status $X^j$.
We assume that the destination can only make a perfect inference of $X^j$, i.e., $\widehat{X}^{j + 1} = X^j$, with a probability of $\varphi \in [0, 1]$ due to the resource and knowledge scarcity.
That is, there exists a probability of $1 - \varphi$ such that $\widehat{X}^{j + 1} \in \mathcal{X} \setminus \{X^j\}$.
%
Different from AoI and AoII, which care only the information freshness at the destination, we adopt AoS to quantify the connection between semantics of the process of interest and inference on the semantic reconstruction of a status update.
More specifically, we define the AoS at the beginning of a time slot $j$ by
\begin{align}\label{AoS}
  c^j = \left(j - \ell^j\right) \cdot \mathds{1}_{\left\{\widehat{X}^j \neq X^j\right\}},
\end{align}
where $\ell^j$ denotes the last time slot when $\widehat{X}^{\ell^j} = X^{\ell^j}$.
Other criteria can be applied to measure the difference between $\widehat{X}^j$ and $X^j$ in the definition of AoS as well, such as the MSE \cite{Sun20}, which is not the focus of this paper.

The AoS dynamics of the SN can be analyzed as in the following two different cases.
\begin{enumerate}
  \item $n^j = 0$: The SN decides not to sample the process status at a time slot $j$.
      In this case, the destination does not receive any new semantic sample from the SN by the end of time slot $j$, which indicates that $\widehat{X}^{j + 1} = \widehat{X}^j$.
      At the beginning of the subsequent time slot $j + 1$, we arrive at $c^{j + 1} = 0$ if the process status switches to $X^{j + 1} = \widehat{X}^{j + 1}$, and otherwise, $c^{j + 1} = \min\{c^j + 1, C\}$.
      Herein, $C$ reflects the staleness of inference by the destination from the received semantic sample.
  \item $n^j = 1$: In this case, the SN samples the process of interest at a time slot $j$.
      From the received semantic sample at the end of time slot $j$, the destination makes an inference $\widehat{X}^{j + 1} = X^j$ with a probability of $\varphi$ or $\widehat{X}^{j + 1} \in \mathcal{X} \setminus \{X^j\}$ with a probability of $1 - \varphi$.
      At the beginning of the next time slot $j + 1$, a) if the process status stays in the same state as at time slot $j$, namely, $X^{j + 1} = X^j$, we have $c^{j + 1} = 0$ when $\widehat{X}^{j + 1} = X^j$, and $c^{j + 1} = \min\{c^j + 1, C\}$ when $\widehat{X}^{j + 1} \in \mathcal{X} \setminus \{X^j\}$;
      and b) if the process status changes to a new state $X^{j + 1} \in \mathcal{X} \setminus \{X^j\}$, $c^{j + 1} $ is then equal to $0$ and $\min\{c^j + 1, C\}$, respectively, conditioned on $\widehat{X}^{j + 1} = X^{j + 1}$ and $\widehat{X}^{j + 1} \in \mathcal{X} \setminus \{X^{j + 1}\}$.
\end{enumerate}

\section{Problem Statement}
\label{prob}

In this section, we first model the joint process status sampling and RS selection in an IRS-aided cooperative relay communication system as an MDP.
The objective of the SN is to maximize the expected discounted utility, which specifies the AoS and energy consumption over the discrete time slots.
After that, we discuss the solution under the online deep RL framework and the corresponding challenges.

\subsection{MDP Formulation}

Since the system involves multiple RSs, the joint process sampling and RS selection is to determine over which RS the SN should transmit the semantic sample to the destination at each time slot.
Under the MDP, the sampling and RS selection actions are adapted to the system states following a control policy.
At the beginning of each time slot $j$, the system state can be encapsulated as $\mathbf{s}^j = (c^j, \mathbf{g}^j, y^j) \in \mathcal{S}$, where $\mathcal{S}$ represents the finite state space\footnote{Although the CSI is generally continuous, we can transform a semi-MDP into a regular discrete MDP with state abstraction \cite{Sutt99}.} and $\mathbf{g}^j = (\mathbf{g}_{(\mathrm{I})}^j, ((g_k^j, \mathbf{g}_{(\mathrm{I}), k}^j, \mathbf{g}_{k, (\mathrm{I})}^j, g_{k, (\mathrm{D})}^j): k \in \mathcal{K}), \mathbf{g}_{(\mathrm{I}, \mathrm{D})}^j)$ represents the CSI profile for the SN.
Let $\pi$ be the stationary control policy of the SN, which maps a system state to a distribution over the actions, namely, $\pi: \mathcal{S} \times \mathcal{A} \rightarrow [0, 1]$, where $\mathcal{A} = \{(0, 0)\} \cup (\{1\} \times \mathcal{K})$ denotes the action space.
By performing an action $\mathbf{a}^j = (n^j, m^j) \in \mathcal{A}$ selected with the probability of $\pi(\mathbf{s}^j, \mathbf{a}^j)$ at each time slot $j$, the system state transits from $\mathbf{s}^j$ to $\mathbf{s}^{j + 1}$ at the beginning of next time slot $j + 1$ with a probability given as
\begin{align}\label{tran_prob}
     \phi\!\left(\mathbf{s}^{j + 1} | \mathbf{s}^j, \mathbf{a}^j\right) =
     \phi\!\left(c^{j + 1} | c^j, n^j\right) \cdot
     \phi\!\left(\mathbf{g}^{j + 1}\right) \cdot
     \phi\!\left(y^{j + 1} | y^j, m^j\right),
\end{align}
and the SN realizes an immediate utility
\begin{align}\label{utility}
    u\!\left(\mathbf{s}^j, \mathbf{a}^j\right) =
    \exp\!\left(- \left(\kappa \cdot c^j + \vartheta \cdot \left(\varrho + p^j \cdot \tau^j_{(1)}\right) \cdot \mathds{1}_{\left\{n^j = 1\right\}}\right)\right),
\end{align}
where $\phi$ denotes the controlled system state transition probability function, while $\kappa$ and $\vartheta$ are the positive weighting factors.
The exponential utility function as in (\ref{utility}) measures the generic satisfaction of the weighted sum of AoS and energy consumption \cite{Fiedler10}.

Executing the stationary control policy $\pi$ across an infinite number of time slots, the state-value function of the SN, which is the expected discounted utility starting from an initial system state $\mathbf{s} =(c, \mathbf{g}, y) \in \mathcal{S}$, can be expressed as
\begin{align}\label{expected_uti}
   V(\mathbf{s}; \pi) =
   (1 - \gamma) \cdot \textsf{E}_{\pi}\!\!\left[\sum_{t = j}^{\infty} (\gamma)^{t - j} \cdot
   u\!\left(\mathbf{s}^t, \mathbf{a}^t\right) | \mathbf{s}^j = \mathbf{s}\right],
\end{align}
where $\gamma \in [0, 1)$ is the discount factor and the expectation $\textsf{E}_{\pi}$ is taken with respect to the probability measure induced by the control policy $\pi$.
As $\gamma$ approaches $1$, the state-value function in (\ref{expected_uti}) also approximates the expected un-discounted utility \cite{Maha96}.
This paper chooses to use the discounted criterion due to the favorable mathematical properties \cite{Tsit07} and the foresight of system uncertainties \cite{Rich98}.
Eventually, the objective of the SN is to find the optimal control policy that maximizes the state-value function.

\subsection{Online DAC Learning and Challenges}
\label{online_DAC}

This section switches to an augmented state-value function $\widetilde{V}(\mathbf{s}; \pi)$ that combines the state-value function $V(\mathbf{s}; \pi)$ and the expected discounted entropy $H(\pi)$ of the control policy $\pi$, $\forall \mathbf{s} \in \mathcal{S}$.
Namely,
\begin{align}\label{aug_state_value}
   \widetilde{V}(\mathbf{s}; \pi) = V(\mathbf{s}; \pi) + \alpha \cdot H(\pi),
\end{align}
where the temperature parameter $\alpha \geq 0$ controls the relative strength and
\begin{align}\label{dis_entropy}
   H(\pi) =
   (1 - \gamma) \cdot \sum_{t = j}^{\infty} (\gamma)^{t - j} \cdot \sum_{\mathbf{a} \in \mathcal{A}}
   \left(- \pi(\mathbf{s}^t, \mathbf{a}) \cdot \ln(\pi(\mathbf{s}^t, \mathbf{a}))\right).
\end{align}
Maximizing the augmented state-value function given by (\ref{aug_state_value}) instead of (\ref{expected_uti}) encourages exploring the action space $\mathcal{A}$ adequately and prohibits the early convergence to sub-optimal control policies \cite{Eyes22}.

By factoring the utility and the entropy at the current time slot in the augmented state-value function (\ref{aug_state_value}), we define the augmented Q-function
\begin{align}\label{aug_Q_value}
     \widetilde{Q}(\mathbf{s}, \mathbf{a}; \pi)
   = (1 - \gamma) \cdot u(\mathbf{s}, \mathbf{a})
   + \gamma \cdot \sum_{\mathbf{s}' \in \mathcal{S}} \phi(\mathbf{s}' | \mathbf{s}, \mathbf{a}) \cdot \widetilde{V}(\mathbf{s}'; \pi),
\end{align}
where $\mathbf{s}' =(c', \mathbf{g}', y')$ denotes the possible subsequent system state after performing an action $\mathbf{a} \in \mathcal{A}$ under the system state $\mathbf{s} \in \mathcal{S}$.
Applying the Bellman equation, we recursively get
\begin{align}\label{aug_state_value_new}
   \widetilde{V}(\mathbf{s}; \pi) =
   \sum_{\mathbf{a} \in \mathcal{A}} \pi(\mathbf{s}, \mathbf{a}) \cdot
   \left(\widetilde{Q}(\mathbf{s}, \mathbf{a}; \pi) - \alpha \cdot (1 - \gamma) \cdot \ln(\pi(\mathbf{s}, \mathbf{a}))\right).
\end{align}
Interleaving the policy evaluation and the policy improvement converges to the optimal stationary control policy \cite{Haar18}.
However, the extremely large state space $\mathcal{S}$ and the dependence on controlled system state transition probability function $\phi$ ask for an DAC scheme to learn the optimal control policy.
To that end, we approximate the optimal control policy and the optimal augmented Q-function using, respectively, a deep actor network $\pi_{\bm\theta}$ and a deep critic network $\widetilde{Q}_{\bm\lambda}$, where $\bm\theta$ and $\bm\lambda$ are the respective deep neural network parameters.

After performing an action $\mathbf{a}^j \in \mathcal{A}$ under the system state $\mathbf{s}^j \in \mathcal{S}$ following the control policy $\pi_{\bm\theta^j}$ at each time slot $j$, the SN trains the deep actor network parameters with the purpose of maximizing the augmented state-value function.
More specifically, the training of the deep actor network follows
\begin{align}\label{actor}
   \bm\theta^{j + 1} \leftarrow \bm\theta^j +
   \beta_{\bm\theta} \cdot \nabla_{\bm\theta^j}\!
   \left(\left(\widetilde{Q}_{\bm\lambda^j}(\mathbf{s}, \mathbf{a}) - \alpha \cdot (1 - \gamma) \cdot \ln(\pi_{\bm\theta^j}(\mathbf{s}, \mathbf{a}))\right) \cdot\ln(\pi_{\bm\theta^j}(\mathbf{s}, \mathbf{a}))\right),
\end{align}
where $\beta_{\bm\theta}$ is the learning rate, while $\bm\theta^j$ and $\bm\lambda^j$ are the parameters of the deep actor network and the deep critic network at time slot $j$.
For the training of the deep critic network, we follow the standard TD method.
In accordance with (\ref{aug_Q_value}) and (\ref{aug_state_value_new}), the on-policy TD error at each time slot $j$ can be mathematically expressed as
\begin{align}\label{TD}
 &   \Delta_{\bm\lambda^j}\!\left(\mathbf{s}^j, \mathbf{a}^j, \mathbf{s}^{j + 1}, \mathbf{a}^{j + 1}\right)         \nonumber\\
 & = (1 - \gamma) \cdot u\!\left(\mathbf{s}^j, \mathbf{a}^j\right) +
     \gamma \cdot \left(\widetilde{Q}_{\bm\lambda^{j, -}}\!\left(\mathbf{s}^{j + 1}, \mathbf{a}^{j + 1}\right) -
     \alpha \cdot (1 - \gamma) \cdot
     \ln\!\left(\pi_{\bm\theta^j}\!\left(\mathbf{s}^{j + 1}, \mathbf{a}^{j + 1}\right)\right)\right)                \nonumber\\
 & - \widetilde{Q}_{\bm\lambda^j}\!\left(\mathbf{s}^j, \mathbf{a}^j\right),
\end{align}
for the state transition from $\mathbf{s}^j$ to $\mathbf{s}^{j + 1} \in \mathcal{S}$, where action $\mathbf{a}^{j + 1} \in \mathcal{A}$ from the control policy $\pi_{\bm\theta^j}$ is performed under $\mathbf{s}^{j + 1}$ and $\bm\lambda^{j, -}$ denotes the deep critic network parameters from a previous time slot before slot $j$.
The SN interacts with the system to adapt the deep critic network parameters such that the TD error is as close to $0$ as possible.
In consequence, the DAC scheme attacks the minimization of
\begin{align}\label{squared_TD}
   \Gamma_{\bm\lambda^j}\!\left(\mathbf{s}^j, \mathbf{a}^j, \mathbf{s}^{j + 1}, \mathbf{a}^{j + 1}\right) =
   \frac{1}{2} \cdot
   \left(\Delta_{\bm\lambda^j}\!\left(\mathbf{s}^j, \mathbf{a}^j, \mathbf{s}^{j + 1}, \mathbf{a}^{j + 1}\right)\right)^2.
\end{align}
After calculating the gradient of (\ref{squared_TD}) with respect to $\bm\lambda^j$, the rule for updating the deep critic network parameters takes the following form
\begin{align}\label{critic}
            \bm\lambda^{j + 1}
 \leftarrow \bm\lambda^j + \beta_{\bm\lambda} \cdot
            \nabla_{\bm\lambda^j} \Gamma_{\bm\lambda^j}\!\left(\mathbf{s}^j, \mathbf{a}^j, \mathbf{s}^{j + 1}, \mathbf{a}^{j + 1}\right),
\end{align}
where $\beta_{\bm\lambda}$ is the learning rate.
Algorithm \ref{algo1} briefly summarizes the procedure of the proposed online DAC scheme.

\begin{algorithm}[!t]
    \caption{Online DAC Scheme for Learning to Optimize Semantics Freshness in IRS-aided Cooperative Relay Communication Systems}
    \label{algo1}
    \begin{algorithmic}[1]
        \STATE initialize the deep actor network parameters $\bm\theta^j$, the deep critic network parameters $\bm\lambda^j$ and $\bm\lambda^{j, -} = \bm\lambda^j$, observe the current system state $\mathbf{s}^j \in \mathcal{S}$, and choose an action $\mathbf{a}^j \in \mathcal{A}$ with probability $\pi_{\bm\theta^j}(\mathbf{s}^j, \mathbf{a}^j)$, for $j = 1$.

        \REPEAT
            \STATE The SN performs the action $\mathbf{a}^j$ and achieves the immediate utility $u(\mathbf{s}^j, \mathbf{a}^j)$.

            \STATE The system transits to the next state $\mathbf{s}^{j + 1} \in \mathcal{S}$.

            \STATE With the observation of $\mathbf{s}^{j + 1}$, the SN chooses an action $\mathbf{a}^{j + 1} \in \mathcal{A}$ with probability $\pi_{\bm\theta^j}(\mathbf{s}^{j + 1}, \mathbf{a}^{j + 1})$.

            \STATE The SN updates the actor network parameters $\bm\theta^{j + 1}$ and the critic network parameters $\bm\lambda^{j + 1}$ according to (\ref{actor}) and (\ref{critic}), respectively.

            \STATE The SN regularly resets the deep critic network parameters by $\bm\lambda^{j + 1, -} = \bm\lambda^{j + 1}$, and otherwise $\bm\lambda^{j + 1, -} = \bm\lambda^{j, -}$.

            \STATE The system time moves to the next slot $j = j + 1$.
        \UNTIL{A predefined stopping condition is satisfied.}
    \end{algorithmic}
\end{algorithm}

\begin{figure}[t]
  \centering
  \subfigure[Online DAC learning.]{\label{onlineDRL}\includegraphics[height=9.63pc]{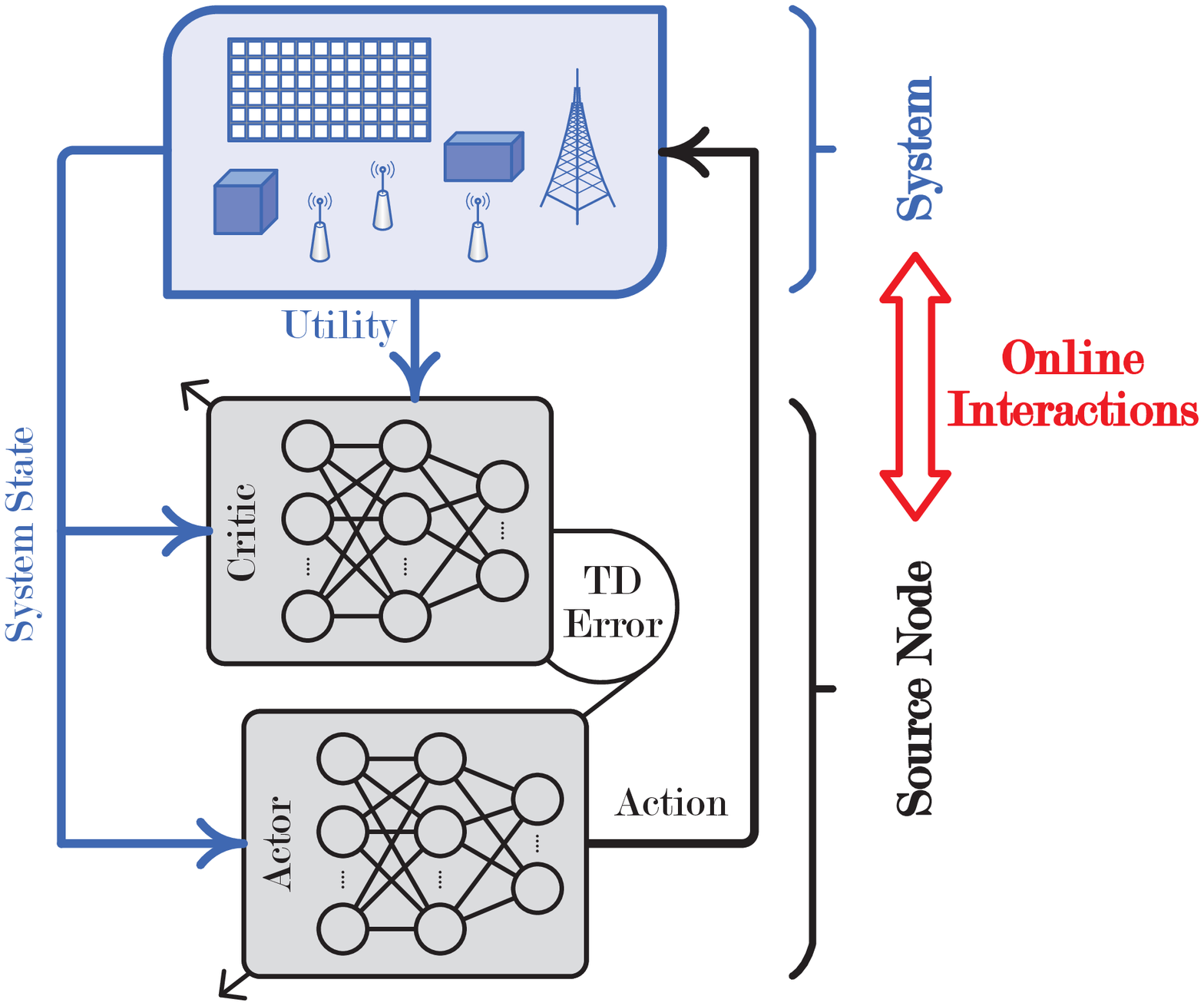}}
  \subfigure[Offline DAC learning.]{\label{offlineDRL}\includegraphics[height=9.63pc]{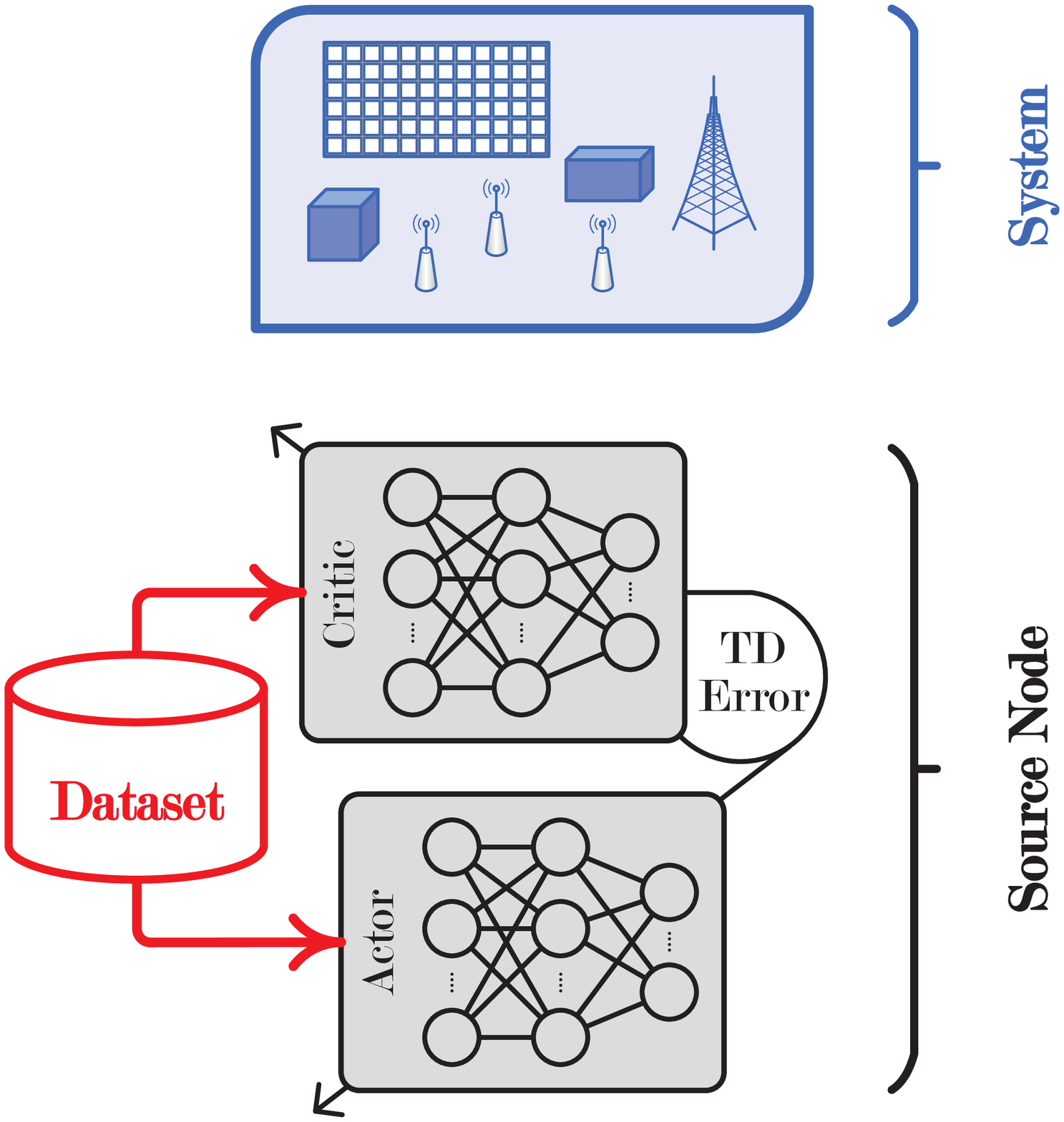}}
  \caption{Implementation comparison between two proposed schemes.}\label{DRL}
\end{figure}

The online implementation of the obtained DAC scheme is displayed in Fig. \ref{onlineDRL}, from which we notice that the learning process alternates over the discrete time slots between optimizing the control policy and collecting new interaction experiences from the policy.
An interaction experience includes the information of current system state, action, immediate utility and subsequent system state.
In other words, the control policy is improved at each time slot relying on the most fresh interaction experience, and meanwhile each new experience comes from the control policy to be optimized \cite{Chen19J, Chen19O}.
Such a ``chicken and egg'' paradox limits the DAC applicability to a real communication system \cite{Dula21}.
Continuous experience acquisition from the online interactions can be expensive and an inappropriate action can lead to painful consequences, particularly during the exploration.
Though the emerging digital twin technology facilitates a virtual simulation to mirror the real communication system, creating a high-fidelity simulator is yet difficult \cite{Wu21}.

\section{Offline Data-Driven Framework}
\label{offline_scheme}

In this section, we shall propose an offline DAC scheme, which aims to learn the target control policy efficiently leveraging a previously collected static dataset.
We designate $\mathcal{D}$ as the dataset consisting of a finite number $|\mathcal{D}|$ of interaction experience tuples, each of which is denoted by $(\mathbf{s}, \mathbf{a}, u(\mathbf{s}, \mathbf{a}), \mathbf{s}') \in \mathcal{D} \subset \mathcal{S} \times \mathcal{A} \times \mathds{R} \times \mathcal{S}$.
Moreover, we let $\pi_{\mathcal{D}}$ denote the empirical control policy that is induced from the dataset $\mathcal{D}$.
%
%
Fig. \ref{offlineDRL} shows the implementation procedure of the offline DAC scheme.
In the following discussions, we slightly abuse the notations from previous Section \ref{prob}.

\subsection{Data-Driven Control Policy Learning}
\label{DD_policy_learning}

The fundamental difference between the online DAC learning and offline data-driven control policy learning is the overestimation from extrapolation of OOD actions \cite{Levine20}.
From the interactions with the system during online DAC learning, the overestimation can be reduced by exploring actions at each time slot, after which the SN updates the control policy with the immediate utility.
Learning from a static dataset offline, there is no opportunity for the SN to interact with the communication system to collect new experience to correct the control policy.
By reformulating the dataset $\mathcal{D}$ as an $|\mathcal{D}|$-time slot trajectory, the online DAC scheme obtained in previous Section \ref{online_DAC} can also be implemented in an offline manner to learn the control policy, akin to behaviour cloning \cite{Wang20}.
However, the performance from the learned control policy highly depends on the dataset quality.
To reduce the extrapolation of OOD actions, this section proposes an offline actor-critic scheme that learns the control policy through lower-bounding the Q-function values.

Given the actor network, the training of the critic network is basically not coupled with the control policy.
By restructuring the state-value function as in (\ref{expected_uti}), we define the Q-function of the SN as
\begin{align}\label{Q_value}
     Q(\mathbf{s}, \mathbf{a}; \pi)
   = (1 - \gamma) \cdot u(\mathbf{s}, \mathbf{a})
   + \gamma \cdot \sum_{\mathbf{s}' \in \mathcal{S}} \phi(\mathbf{s}' | \mathbf{s}, \mathbf{a}) \cdot V(\mathbf{s}'; \pi),
\end{align}
which describes the expected discounted utility for performing an action $\mathbf{a} \in \mathcal{A}$ under a system state $\mathbf{s} \in \mathcal{S}$ and following the stationary control policy $\pi$ thereafter.
In turn, we get $V(\mathbf{s}; \pi) = \sum_{\mathbf{a} \in \mathcal{A}} \pi(\mathbf{s}, \mathbf{a}) \cdot Q(\mathbf{s}, \mathbf{a}; \pi)$, with which the Q-function can be reexpressed as the following Bellman equation
\begin{align}\label{Q_value2}
     Q(\mathbf{s}, \mathbf{a}; \pi)
   = (1 - \gamma) \cdot u(\mathbf{s}, \mathbf{a})
   + \gamma \cdot \sum_{\mathbf{s}' \in \mathcal{S}} \phi(\mathbf{s}' | \mathbf{s}, \mathbf{a}) \cdot
     \sum_{\mathbf{a}' \in \mathcal{A}} \pi(\mathbf{s}', \mathbf{a}') \cdot Q(\mathbf{s}', \mathbf{a}'; \pi).
\end{align}
Given the dataset $\mathcal{D}$ under the empirical control policy $\pi_{\mathcal{D}}$ in the offline settings, an estimated control policy is evaluated to determine the Q-function values.
Since $\mathcal{D}$ may not include all possible interaction experiences in $\mathcal{S} \times \mathcal{A} \times \mathds{R} \times \mathcal{S}$, the SN trains the critic network by iterating the estimated Q-function in order to minimize the MSE of the Bellman equation (\ref{Q_value2}), namely, the loss function at each iteration $j$ given by
\begin{align}\label{expected_TD}
 & l\!\left(\hat{Q}; \hat{\pi}^j, \hat{Q}^j, \mathcal{O}^j\right) =                                                                 \\
 & \frac{1}{2} \cdot \textsf{E}_{\{(\mathbf{s}, \mathbf{a}, u(\mathbf{s}, \mathbf{a}), \mathbf{s}') \in \mathcal{O}^j\}}\!\!
   \left[\left(\hat{Q}(\mathbf{s}, \mathbf{a}) - \left((1 - \gamma) \cdot u(\mathbf{s}, \mathbf{a}) +
   \gamma \cdot \sum_{\mathbf{a}' \in \mathcal{A}} \hat{\pi}^j(\mathbf{s}', \mathbf{a}') \cdot \hat{Q}^j(\mathbf{s}', \mathbf{a}')\right)
   \right)^2\right],                                                                                                                \nonumber
\end{align}
where $\mathcal{O}^j \subset \mathcal{D}$ is a random mini-batch from $\mathcal{D}$ and is of size $|\mathcal{O}^j| = O$, while $\hat{Q}^j$ and $\hat{\pi}^j$ are estimates of the optimal Q-function and the optimal control policy, respectively.
With the output of the critic network at an iteration $j$, the estimated control policy $\hat{\pi}$ is improved by training the actor network to maximize
\begin{align}\label{estimated_state_value}
    f\!\left(\hat{\pi}; \hat{Q}^j, \mathcal{O}^j\right) =
    \textsf{E}_{\{\mathbf{s}: (\mathbf{s}, \mathbf{a}, u(\mathbf{s}, \mathbf{a}), \mathbf{s}') \in \mathcal{O}^j\}}\!\!
    \left[\sum_{\mathbf{a}' \in \mathcal{A}} \hat{\pi}(\mathbf{s}, \mathbf{a}') \cdot \hat{Q}^j(\mathbf{s}, \mathbf{a}')\right].
\end{align}
It can be easily found that the above training of critic and actor networks does not skip the challenge of action distribution shift \cite{Kumar20}.
On one hand, as in (\ref{expected_TD}), the Q-function is only trained for a system state and an action that appear in each interaction experience tuple from the dataset $\mathcal{D}$, but the value of the subsequent system state is calculated as the sum of Q-function values weighted by the estimated control policy over all actions.
On the other hand, the control policy is trained to maximize the mean state-value in (\ref{estimated_state_value}), which biases the OOD actions with inaccurate high Q-function values.

To steer the offline control policy learning away from the OOD actions, we pose the penalties below on the estimated Q-function values during the critic network training.
\begin{enumerate}
  \item The estimated Q-function, with which the estimated control policy improves upon the empirical control policy \cite{Abdolmaleki18}, is minimized together with (\ref{expected_TD}).
  \item The Q-function values are regularized to rank the actions appearing in the dataset higher than the OOD actions, which can be interpreted as that the critic network training adjusts the estimated Q-function of an OOD action if the value is larger than that of an action from the dataset by a certain margin \cite{Su20, Lin20}.
\end{enumerate}
Correspondingly, we recast the loss function given by (\ref{expected_TD}) into an augmented loss function as
\begin{align}\label{aug_expected_TD}
 &   L\!\left(\hat{Q}; \hat{\pi}^j, \hat{Q}^j, \mathcal{O}^j\right)\\
 & = \frac{1}{2} \cdot \textsf{E}_{\{(\mathbf{s}, \mathbf{a}, u(\mathbf{s}, \mathbf{a}), \mathbf{s}') \in \mathcal{O}^j\}}\!\!
     \left[\left(\hat{Q}(\mathbf{s}, \mathbf{a}) - \left((1 - \gamma) \cdot u(\mathbf{s}, \mathbf{a}) +
     \gamma \cdot \sum_{\mathbf{a}' \in \mathcal{A}} \hat{\pi}^j(\mathbf{s}', \mathbf{a}') \cdot \hat{Q}^j(\mathbf{s}', \mathbf{a}')\right)
     \right)^2\right]                                                                                                       \nonumber\\
 & + \rho \cdot
     \textsf{E}_{\{(\mathbf{s}, \mathbf{a}): (\mathbf{s}, \mathbf{a}, u(\mathbf{s}, \mathbf{a}), \mathbf{s}') \in \mathcal{O}^j\}}\!\!
     \left[\sum_{\mathbf{a}' \in \mathcal{A}} \omega(\mathbf{s}, \mathbf{a}') \cdot \hat{Q}(\mathbf{s}, \mathbf{a}') +
     \sum_{\mathbf{a}' \in \mathcal{A} \setminus \{\mathbf{a}\}} \max\!\left\{0, \nu + \hat{Q}(\mathbf{s}, \mathbf{a}') - \hat{Q}(\mathbf{s}, \mathbf{a})\right\}\right],                                                                                            \nonumber
\end{align}
where $\omega$ is a control policy that sharpens the empirical control policy $\pi_{\mathcal{D}}$ and will be discussed in details in Section \ref{P_offline_DAC}, while the constant $\rho$ trades off the penalty degree and $\nu$ denotes the positive margin.
It is worth mentioning that the support of $\omega$ caters to $\mathrm{supp}(\omega) \subset \mathrm{supp}(\pi_{\mathcal{D}})$ \cite{Rezaeifar22}.
For an estimated control policy $\hat{\pi}$, the expected discounted entropy in (\ref{dis_entropy}) can be rewritten as the sum of infinite geometric series.
That is, we have
\begin{align}\label{re_dis_entropy}
   H(\hat{\pi}) = - \sum_{\mathbf{s} \in \mathcal{S}} \sum_{\mathbf{a} \in \mathcal{A}}
   \hat{\pi}(\mathbf{s}, \mathbf{a}) \cdot \ln(\hat{\pi}(\mathbf{s}, \mathbf{a})).
\end{align}
With a dataset $\mathcal{D}$ with a limited number of interaction experience tuples, the empirical control policy $\pi_{\mathcal{D}}$ tends to be deterministic, particularly when the state space is exceptionally large (as in the numerical experiments).
It is obvious that a low-quality dataset hurts the actor network training.
Therefore, we replace the objective in (\ref{estimated_state_value}) with
\begin{align}\label{estimated_augmented_state_value}
  F\!\left(\hat{\pi}; \hat{Q}^j, \mathcal{O}^j\right) =
  \textsf{E}_{\{\mathbf{s}: (\mathbf{s}, \mathbf{a}, u(\mathbf{s}, \mathbf{a}), \mathbf{s}') \in \mathcal{O}^j\}}\!\!
  \left[\sum_{\mathbf{a}' \in \mathcal{A}} \hat{\pi}(\mathbf{s}, \mathbf{a}') \cdot \left(\hat{Q}^j(\mathbf{s}, \mathbf{a}') -
  \alpha \cdot \ln(\hat{\pi}(\mathbf{s}, \mathbf{a}'))\right)\right],
\end{align}
where the entropy term based on (\ref{re_dis_entropy}) results in a more stochastic control policy.

\subsection{Theoretical Analysis of Bounded Q-Function Estimation}

In the previous Section \ref{DD_policy_learning}, the augmented loss function given by (\ref{aug_expected_TD}) includes two penalty terms.
The minimization of the first penalty term lower bounds the estimated Q-function \cite{Donoghue21}.
As confirmed by Lemma 1, the second penalty term further improves the lower bound of the estimated Q-function of OOD actions.

\emph{Lemma 1.}
A lower bound of the estimated Q-function of OOD actions can be approximately optimized by the minimization of
\begin{align}\label{op_prob}
  z(\mathbf{s}, \mathbf{a}) = \sum_{\mathbf{a}' \in \mathcal{A} \setminus \{\mathbf{a}\}} \max\!\left\{0, \nu + \hat{Q}(\mathbf{s}, \mathbf{a}') - \hat{Q}(\mathbf{s}, \mathbf{a})\right\},
\end{align}
for $(\mathbf{s}, \mathbf{a})$ appearing in each interaction experience tuple from the static dataset $\mathcal{D}$.

\emph{Proof:}
Using the ranking policy gradient theorem in \cite{Lin20}, maximizing the expected discounted utility performance is equivalent to optimizing the ranking control policy, which can be given by
\begin{align}\label{ranking_policy}
  \varpi(\mathbf{s}, \mathbf{a}) = \prod_{\mathbf{a}' \in \mathcal{A} \setminus \{\mathbf{a}\}}
  \frac{\exp\!\left(\hat{Q}(\mathbf{s}, \mathbf{a}) - \hat{Q}(\mathbf{s}, \mathbf{a}')\right)}
       {1 + \exp\!\left(\hat{Q}(\mathbf{s}, \mathbf{a}) - \hat{Q}(\mathbf{s}, \mathbf{a}')\right)},
\end{align}
for each $(\mathbf{s}, \mathbf{a}) \in \mathcal{S} \times \mathcal{A}$.
Given the static dataset $\mathcal{D}$ of a finite number of interaction experience tuples, the expected discounted utility maximization is essentially to maximize the log-likelihood of each existing $(\mathbf{s}, \mathbf{a})$, following which we have
\begin{align}
      \max_{\hat{Q}} \ln(\varpi(\mathbf{s}, \mathbf{a}))
  & = \max_{\hat{Q}} \ln\!\left(\prod_{\mathbf{a}' \in \mathcal{A} \setminus \{\mathbf{a}\}}
      \frac{\exp\!\left(\hat{Q}(\mathbf{s}, \mathbf{a}) - \hat{Q}(\mathbf{s}, \mathbf{a}')\right)}
           {1 + \exp\!\left(\hat{Q}(\mathbf{s}, \mathbf{a}) - \hat{Q}(\mathbf{s}, \mathbf{a}')\right)}\right)       \nonumber\\
  & = \max_{\hat{Q}} \sum_{\mathbf{a}' \in \mathcal{A} \setminus \{\mathbf{a}\}}
      \left(\left(\hat{Q}(\mathbf{s}, \mathbf{a}) - \hat{Q}(\mathbf{s}, \mathbf{a}')\right)\right. -                \nonumber\\
  &   \qquad\qquad\qquad\quad\left.\ln\!\left(1 + \exp\!\left(\hat{Q}(\mathbf{s}, \mathbf{a}) -
      \hat{Q}(\mathbf{s}, \mathbf{a}')\right)\right)\right)                                                         \nonumber\\
  & \approx \min_{\hat{Q}} \frac{1}{2} \cdot \sum_{\mathbf{a}' \in \mathcal{A} \setminus \{\mathbf{a}\}}
      \left(\ln(4) + \hat{Q}(\mathbf{s}, \mathbf{a}') - \hat{Q}(\mathbf{s}, \mathbf{a})\right)                      \label{op_prob1}\\
  & \Leftrightarrow \min_{\hat{Q}} z(\mathbf{s}, \mathbf{a}).                                                       \nonumber
\end{align}
From the Q-function definition, we have $|\hat{Q}(\mathbf{s}, \mathbf{a}) - \hat{Q}(\mathbf{s}, \mathbf{a}')| \leq u_{(\max)} \leq 1$, where $u_{(\max)} = \max_{(\mathbf{s}, \mathbf{a}) \in \mathcal{S} \times \mathcal{A}} u(\mathbf{s}, \mathbf{a})$.
It is easy to find that the ranking policy from (\ref{op_prob1}) is consistent with the policy from minimizing (\ref{op_prob}).
Therefore, we are able to take the minimization of (\ref{op_prob}) as a surrogate\footnote{The reason of using (\ref{op_prob}) rather than (\ref{op_prob1}) in the augmented loss function is to stabilize the training process. In the numerical experiments, the estimated Q-function parameters are randomly initialized.} of (\ref{op_prob1}), which completes the proof.
\hfill $\square$

Next, we provide the theoretical support that the minimization of the augmented loss function (\ref{aug_expected_TD}) guarantees the lower-bound of the estimated Q-function.
We let $\mathcal{Q}$ be the space of the bounded real-valued functions over $\mathcal{S} \times \mathcal{A}$.
For the given target stationary control policy $\pi$, we define the Bellman operator $\mathcal{T}^{\pi}$ by
\begin{align}\label{Bellman_operator}
    \mathcal{T}^{\pi}Q(\mathbf{s}, \mathbf{a}; \pi) = (1 - \gamma) \cdot u(\mathbf{s}, \mathbf{a}) +
    \gamma \cdot \sum_{\mathbf{s}' \in \mathcal{S}} \phi(\mathbf{s}' | \mathbf{s}, \mathbf{a}) \cdot
    \sum_{\mathbf{a}' \in \mathcal{A}} \pi(\mathbf{s}', \mathbf{a}') \cdot Q(\mathbf{s}', \mathbf{a}'; \pi),
\end{align}
$\forall \mathbf{s} \in \mathcal{S}$ and $\forall \mathbf{a} \in \mathcal{A}$, which is a mapping $\mathcal{T}^{\pi}: \mathcal{Q} \rightarrow \mathcal{Q}$.
In our offline settings without any interactions with the communication system, (\ref{aug_expected_TD}) uses an estimated Bellman operator $\mathcal{T}^{\hat{\pi}^j}$, where $\hat{\pi}^j$ is an estimate of $\pi$ based on the dataset $\mathcal{D}$ at each iteration $j$.
Following the martingale concentration inequality \cite{Min22},
\begin{align}\label{concentration_property}
  \left|\mathcal{T}^{\pi}Q(\mathbf{s}, \mathbf{a}; \pi) - \mathcal{T}^{\pi}\hat{Q}(\mathbf{s}, \mathbf{a})\right| \leq
  \frac{\psi}{\sqrt{\max\{1, |\mathcal{D}(\mathbf{s}, \mathbf{a})|\}}}.
\end{align}
holds with the probability of $1 - \epsilon$ for an $\epsilon \in (0, 1)$, where $\psi$ is a constant that depends on the statistics of utility realizations in $\mathcal{D}$, while $\mathcal{D}(\mathbf{s}, \mathbf{a}) \subset \mathcal{D}$ denotes the subset of interaction experience tuples including $(\mathbf{s}, \mathbf{a})$.
In the analysis that follows, let $\bm\omega^\pi = [\omega^\pi_{\mathbf{s}}: \mathbf{s} \in \mathcal{S}]_{|\mathcal{S}| \times 1}$ and $\mathbf{d}^\pi = [d^\pi_{\mathbf{s}}: \mathbf{s} \in \mathcal{S}]_{|\mathcal{S}| \times 1}$ be two column vectors with $\omega^\pi_{\mathbf{s}} = \sum_{\mathbf{a} \in \mathcal{A}} (\pi(\mathbf{s}, \mathbf{a}) \cdot (\omega(\mathbf{s}, \mathbf{a}) + 1)) /$ $\pi_{\mathcal{D}}(\mathbf{s}, \mathbf{a})$ and $d^\pi_{\mathbf{s}} = \psi \cdot \sum_{\mathbf{a} \in \mathcal{A}} \pi(\mathbf{s}, \mathbf{a}) / \sqrt{\max\{1, |\mathcal{D}(\mathbf{s}, \mathbf{a})|\}}$.
As one of the major results from this paper, we have Theorem 2 stated as below.

\emph{Theorem 2.}
There exists an $\epsilon \in (0, 1)$ such that with the probability $1 - \epsilon$, the state-value function under the estimated Q-function $\hat{Q}$ from minimizing the augmented loss function as in (\ref{aug_expected_TD}) satisfies\footnote{For the analysis convenience, the system states, which do not appear in the dateset $\mathcal{D}$, are not excluded from the estimated state-value function. The theoretical analysis still holds by letting the estimated state-values of such states equal to the state-values.}
\begin{align}\label{lower_bound}
        \hat{\mathbf{V}}^{\pi}
   \leq \mathbf{V}(\pi) - \left(\mathbf{I} - \gamma \cdot \bm\Phi^{\pi}\right)^{- 1} \cdot \left(\rho \cdot
        \bm\omega^\pi - \mathbf{d}^\pi\right),
\end{align}
wherein $\hat{\mathbf{V}}^{\pi} = [\hat{V}^{\pi}(\mathbf{s}): \mathbf{s} \in \mathcal{S}]_{|\mathcal{S}| \times 1}$ with each $\hat{V}(\mathbf{s}) = \sum_{\mathbf{a} \in \mathcal{A}} \pi(\mathbf{s}, \mathbf{a}) \cdot \hat{Q}(\mathbf{s}, \mathbf{a})$, $\mathbf{V}(\pi) = [V(\mathbf{s}; \pi): \mathbf{s} \in \mathcal{S}]_{|\mathcal{S}| \times 1}$, $\mathbf{I}$ denotes an $|\mathcal{S}| \times |\mathcal{S}|$ identity matrix, and $\bm\Phi^{\pi}$ is an $|\mathcal{S}| \times |\mathcal{S}|$ matrix with each entry at the position $(e_\mathbf{s}, e_{\mathbf{s}'})$ ($1 \leq e_\mathbf{s}, e_{\mathbf{s}'} \leq |\mathcal{S}|$) given by $\bm\Phi^{\pi}_{e_\mathbf{s}, e_{\mathbf{s}'}} = \sum_{\mathbf{a} \in \mathcal{A}} \pi(\mathbf{s}, \mathbf{a}) \cdot \phi(\mathbf{s}' | \mathbf{s}, \mathbf{a})$.
If
\begin{align}\label{penalty_constant}
   \rho \geq
   \max_{(\mathbf{s}, \mathbf{a}) \in \mathcal{S} \times \mathcal{A}}
   \frac{\psi}{\sqrt{\max\{1, |\mathcal{D}(\mathbf{s}, \mathbf{a})|\}}} \cdot
   \frac{\pi_{\mathcal{D}}(\mathbf{s}, \mathbf{a})}{\omega(\mathbf{s}, \mathbf{a}) + 1},
\end{align}
then for any system state $\mathbf{s}$ in an interaction experience tuple from the dataset $\mathcal{D}$, $\hat{V}(\mathbf{s}) \leq V(\mathbf{s}; \pi)$.

\emph{Proof:}
By calculating the derivative of the augmented loss function $L(\hat{Q}; \hat{\pi}^j, \mathcal{O}^j)$ given by (\ref{aug_expected_TD}) with respect to $\hat{Q}(\mathbf{s}, \mathbf{a})$ and setting it to $0$, we attain
\begin{align}
  \hat{Q}(\mathbf{s}, \mathbf{a}) = \mathcal{T}^{\hat{\pi}^j} \hat{Q}^j(\mathbf{s}, \mathbf{a}) -
  \rho \cdot \frac{\omega(\mathbf{s}, \mathbf{a}) + 1}{\pi_{\mathcal{D}}(\mathbf{s}, \mathbf{a})},
\end{align}
for any $(\mathbf{s}, \mathbf{a})$ in an interaction experience tuple from $\mathcal{D}$, where the derivation of the second term at the right-hand-side is based on the marginal system state distribution under the empirical control policy $\pi_{\mathcal{D}}$ \cite{Sharma22}.
It can be noted that the minimization of augmented loss function leads to the upper-bounded estimated Q-function, namely,
\begin{align}
    \hat{Q}(\mathbf{s}, \mathbf{a}) \leq \mathcal{T}^{\hat{\pi}^j} \hat{Q}^j(\mathbf{s}, \mathbf{a}) = \hat{Q}^j(\mathbf{s}, \mathbf{a}).
\end{align}
Recall the concentration property as in (\ref{concentration_property}), we hence obtain the following
\begin{align}
        \hat{Q}(\mathbf{s}, \mathbf{a})
 & =    \mathcal{T}^{\hat{\pi}} \hat{Q}(\mathbf{s}, \mathbf{a})                                             \nonumber\\
 & \leq \mathcal{T}^{\pi} \hat{Q}(\mathbf{s}, \mathbf{a}) -
        \rho \cdot \frac{\omega(\mathbf{s}, \mathbf{a}) + 1}{\pi_{\mathcal{D}}(\mathbf{s}, \mathbf{a})}
   +    \frac{\psi}{\sqrt{\max\{1, |\mathcal{D}(\mathbf{s}, \mathbf{a})|\}}},
\end{align}
with the probability $1 - \epsilon$, which indicates that the estimated state-value $\hat{V}(\mathbf{s})$ from the estimated Q-function $\hat{Q}(\mathbf{s}, \mathbf{a})$ and the control policy $\pi$ fulfills
\begin{align}\label{proof2_01}
        \left(\mathbf{I} - \gamma \cdot \bm\Phi^{\pi}\right) \cdot \hat{\mathbf{V}}^{\pi}
   \leq (1 - \gamma) \cdot \mathbf{u} - \rho \cdot \bm\omega^\pi + \mathbf{d}^\pi,
\end{align}
where $\mathbf{u} = [u^\pi_{\mathbf{s}}: \mathbf{s} \in \mathcal{S}]_{|\mathcal{S}| \times 1}$ with each $u^\pi_{\mathbf{s}} = \sum_{\mathbf{a} \in \mathcal{A}} \pi(\mathbf{s}, \mathbf{a}) \cdot u(\mathbf{s}, \mathbf{a})$.
For $\gamma < 1$, the spectral radius of $\gamma \cdot \bm\Phi^{\pi}$ is smaller than $1$, hence the inverse of $(\mathbf{I} - \gamma \cdot \bm\Phi^{\pi})$ exists.
By multiplying $(\mathbf{I} - \gamma \cdot \bm\Phi^{\pi})^{- 1}$ with both sides of (\ref{proof2_01}), we acquire (\ref{lower_bound}) by taking into account that $\mathbf{V}(\pi) = (1 - \gamma) \cdot (\mathbf{I} - \gamma \cdot \bm\Phi^{\pi})^{- 1} \cdot \mathbf{u}$.
Consequently, the penalty constant $\rho$ chosen according to (\ref{penalty_constant}) prevents the extrapolation error of the estimated Q-function.
This completes the proof of Theorem 2.
\hfill $\square$

\subsection{Practical Offline DAC Scheme}
\label{P_offline_DAC}

Following \cite{Wang20, Nachum17}, the optimum $\omega^*$ in the augmented loss function (\ref{aug_expected_TD}) can be realized by maximizing
\begin{align}\label{KL}
     b(\omega)
   = \textsf{E}_{\{\mathbf{s}: (\mathbf{s}, \mathbf{a}, u(\mathbf{s}, \mathbf{a}), \mathbf{s}') \in \mathcal{D}\}}\!\!
     \left[\sum_{\mathbf{a}' \in \mathcal{A}} \omega(\mathbf{s}, \mathbf{a}') \cdot \hat{Q}(\mathbf{s}, \mathbf{a}')\right]
   - \textsf{KL}(\omega, \pi_{\mathcal{D}}),
\end{align}
where $\textsf{KL}(\omega, \pi_{\mathcal{D}})$ means the Kullback–Leibler divergence between $\omega$ and $\pi_{\mathcal{D}}$.
Setting the derivative of $b(\omega)$ with respect to $\hat{Q}(\mathbf{s}, \mathbf{a}')$ to $0$ yields
\begin{align}\label{KL_op}
  \omega^*(\mathbf{s}, \mathbf{a}') =
  \pi_{\mathcal{D}}(\mathbf{s}, \mathbf{a}') \cdot \exp\!\left(\hat{Q}(\mathbf{s}, \mathbf{a}') - 1\right),
\end{align}
for the system state $\mathbf{s}$ in an interaction experience tuple from the dataset $\mathcal{D}$ and each $\mathbf{a}' \in \mathcal{A}$.
By substituting (\ref{KL_op}) back into (\ref{KL}), we thus can rewrite (\ref{aug_expected_TD}) as
\begin{align}\label{aug_expected_TD_app}
 &   L\!\left(\hat{Q}; \hat{\pi}^j, \mathcal{O}^j\right)\\
 & = \frac{1}{2} \cdot \textsf{E}_{\{(\mathbf{s}, \mathbf{a}, u(\mathbf{s}, \mathbf{a}), \mathbf{s}') \in \mathcal{O}^j\}}\!\!
     \left[\left(\hat{Q}(\mathbf{s}, \mathbf{a}) - \left((1 - \gamma) \cdot u(\mathbf{s}, \mathbf{a}) +
     \gamma \cdot \sum_{\mathbf{a}' \in \mathcal{A}} \hat{\pi}^j(\mathbf{s}', \mathbf{a}') \cdot \hat{Q}^j(\mathbf{s}', \mathbf{a}')\right)
     \right)^2\right] \nonumber\\
 & + \rho \cdot
     \textsf{E}_{\{(\mathbf{s}, \mathbf{a}): (\mathbf{s}, \mathbf{a}, u(\mathbf{s}, \mathbf{a}), \mathbf{s}') \in \mathcal{O}^j\}}\!\!
     \left[\ln\!\left(\sum_{\mathbf{a}' \in \mathcal{A}} \exp\!\left(\hat{Q}(\mathbf{s}, \mathbf{a}')\right)\right) +
     \sum_{\mathbf{a}' \in \mathcal{A} \setminus \{\mathbf{a}\}} \max\!\left\{0, \nu + \hat{Q}(\mathbf{s}, \mathbf{a}') - \hat{Q}(\mathbf{s}, \mathbf{a})\right\}\right],\nonumber
\end{align}
by replacing the first penalty with a softmax value $\ln(\sum_{\mathbf{a}' \in \mathcal{A}} \exp(\hat{Q}(\mathbf{s}, \mathbf{a}')))$ \cite{Haarnoja17}.
To address the challenge of an extremely large state space, we employ two deep neural networks $\hat{\pi}_{\bm\theta^j}$ and $\hat{Q}_{\bm\lambda^j}$ to model the estimated control policy $\hat{\pi}^j$ and the estimated Q-function $\hat{Q}^j$ at each iteration $j$, as in the online settings.
To be specific, the updating rules for the deep actor network and the deep critic network parameters are given by
\begin{align}\label{off_actor}
   \bm\theta^{j + 1} \leftarrow \bm\theta^j + \beta_{\bm\theta} \cdot \nabla_{\bm\theta^j} F\!\left(\hat{\pi}_{\bm\theta^j}; \hat{Q}_{\bm\lambda^j}, \mathcal{O}^j\right),
\end{align}
and
\begin{align}\label{off_critic}
               \bm\lambda^{j + 1}
    \leftarrow \bm\lambda^j - \beta_{\bm\lambda} \cdot \nabla_{\bm\lambda^j} L\!\left(\hat{Q}_{\bm\lambda^j}; \hat{\pi}_{\bm\theta^j}, \hat{Q}_{\bm\lambda^{j, -}}, \mathcal{O}^j\right),
\end{align}
respectively, where we choose $\bm\lambda^{j, -}$ to denote the deep critic network parameters from a previous iteration before iteration $j$ and is regularly reset.
Algorithm \ref{algo2} summarizes the implementation procedure of our proposed offline DAC scheme.

\begin{algorithm}[!t]
    \caption{Offline DAC Scheme for Semantics Freshness Optimization in IRS-aided Cooperative Relay Communication Systems}
    \label{algo2}
    \begin{algorithmic}[1]
        \STATE initialize the deep actor network parameters $\bm\theta^j$ as well as the deep critic network parameters $\bm\lambda^j$, for $j = 1$.

        \REPEAT
            \STATE Randomly sample a mini-batch $\mathcal{O}^j \subset \mathcal{D}$ of interaction experience tuples.

            \STATE Update the deep actor network parameters $\bm\theta^{j + 1}$ according to (\ref{off_actor}).

            \STATE Update the deep critic network parameters $\bm\lambda^{j + 1}$ according to (\ref{off_critic}).

            \STATE Regularly reset the deep critic network parameters with $\bm\lambda^{j + 1, -} = \bm\lambda^{j + 1}$, and otherwise $\bm\lambda^{j + 1, -} = \bm\lambda^{j, -}$.

            \STATE Set the iteration index $j = j + 1$.
        \UNTIL{A predefined stopping condition is satisfied.}
    \end{algorithmic}
\end{algorithm}

\section{Numerical Experiments}
\label{simu}

In this section, we numerically evaluate the proposed offline DAC scheme by conducting a series of experiments with TensorFlow.

\subsection{Experimental Configurations and Datasets}

We set up an IRS-aided cooperative relay communication system with $K = 5$ RSs, which is similar to the scenario in \cite{Wang13}.
In the system, the IRS is deployed to have line-of-sight channels to the SN/RSs/destination.
Moreover, there are non-line-of-sight channels between the SN and the RSs as well as between the RSs and the destination due to the obstacles.
The channel gains over the discrete time slots are hence modelled following \cite{Bjor20} and taking into account the environmental disturbances.
At each time slot, the status of the process of interest at the SN is assumed to be in one of $|\mathcal{X}| = 9$ states, for which we set the probability of remaining in the same state during the next time slot as $\chi$.
Then for experimental purpose, the probability of transitioning to another different state is $(1 - \chi) / (|\mathcal{X}| - 1)$.
For both of the online DAC and the offline DAC schemes, the deep actor and the deep critic networks are designed to be with one hidden layer, which contains $64$ neurons and uses ReLU as the activation function \cite{Nair10}.
As for the output layer, the deep actor network chooses Softmax as the activation function, while the deep critic network selects a linear output layer \cite{Goodfellow16}.
Adam is kept as the optimizer throughout all experiments \cite{King15}.
Other parameter values are listed in Table \ref{tabl1}.

\begin{table}[t]
  \caption{Parameter values in experiments.}\label{tabl1}
        \begin{center}
        \begin{tabular}{|c|c||c|c|}
              \hline
              Parameter       & Value                     & Parameter     & Value                               \\\hline
              \hline
              $\zeta$         & $1$                       & $P$             & $30$ dBm                          \\\hline
              $w$             & $10$ MHz                  & $\kappa$        & $5 \cdot 10^{- 2}$                \\\hline
              $C$             & $30$                      & $\vartheta$     & $1$                               \\\hline
              $\upsilon$      & $6.2 \cdot 10^6$ bits     & $\sigma^2$      & $-174$ dBm/Hz                     \\\hline
              $\varrho$       & $0.01$ Joule              & $\gamma$        & $0.9$                             \\\hline
              $\tau$          & $0.1$ seconds             & $\alpha$        & $10^{-4}$                         \\\hline
              $\delta$        & $10^{- 3}$ seconds        & $\rho$          & $5 \cdot 10^{-4}$                 \\\hline
              $P_k$           & $30$ dBm, $\forall k$     & $\nu$           & $1$                               \\
              \hline
        \end{tabular}
        \end{center}
\end{table}

In addition to the online DAC scheme, we compare the proposed offline DAC scheme with the three baselines as well, namely, the A2C scheme \cite{Mnih16}, the Random scheme and the CQL scheme \cite{Kumar20}.
Implementing the Random scheme, the SN applies a uniform probability distribution over the sampling and RS selection actions across the time horizon.
All datasets in experiments are generated from both the A2C and the Random schemes, and are categorized into the respective Expert Data and Random Data.
For each dataset, we collect $|\mathcal{D}| = 2 \cdot 10^5$ interaction experience tuples.
The size of each mini-batch is set to be $O = 5 \cdot 10^3$ during the training of our proposed offline DAC scheme.

\subsection{Results and Discussions}

\subsubsection{Convergence Validation}

\begin{figure}[t]
  \centering
  \includegraphics[width=19pc]{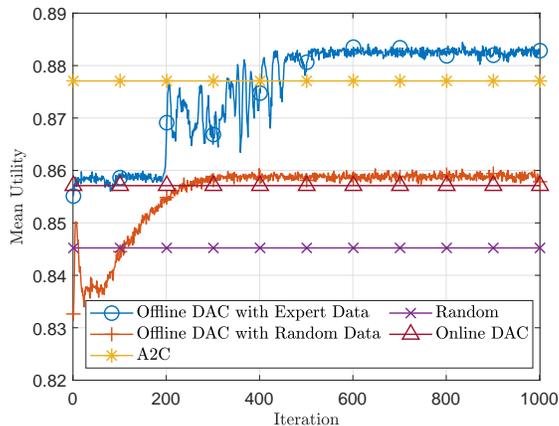}
  \caption{Illustration of convergence speed of the proposed offline DAC scheme in terms of mean utility: $I = 75$, $\varphi = 0.5$ and $\chi = 0.5$.}
  \label{simu01}
\end{figure}

We first evaluate the convergence speed of training the proposed offline DAC scheme using not only the Expert Data but also the Random Data.
In the experiment, we assume an IRS with $I = 75$ reflecting elements, while the accurate inference construction probability at the destination and the state-remaining probability for the process at the SN are set to be $\varphi = 0.5$ and $\chi = 0.5$, respectively.
We plot the variations in the mean utility during the training of our proposed offline DAC scheme in Fig. \ref{simu01}, which shows the mean utility performance of the A2C, Random and online DAC schemes as well.
Each point on the curves of the proposed offline DAC scheme corresponds to the mean of $10^5$ utility realizations from online testing the deep actor network parameters, which are offline trained at each iteration.
The curves clearly tell that the offline training converges within $600$ iterations.
Besides, the converged offline DAC schemes trained using Expert Data and Random Data outperform the respective A2C and Random schemes.
This is attributed to the exploration/exploitation tradeoff during the online A2C learning.
It is interesting to see that the proposed offline DAC scheme even trained with Random Data achieves better mean utility performance than the proposed online DAC scheme.
Unsurprisingly, the mean utility performance from the online DAC scheme is deteriorated compared to the A2C scheme.
The reason is that the online DAC scheme performs on-policy learning (as in (\ref{TD})) to learn the near-optimal control policy in a more conservative way than the off-policy A2C scheme, which tends to be aggressive and learns directly the optimal control policy \cite{Mnih16}.

\subsubsection{Performance Comparison With Baselines}

\begin{figure}[t]
    \centering
    \includegraphics[width=19pc]{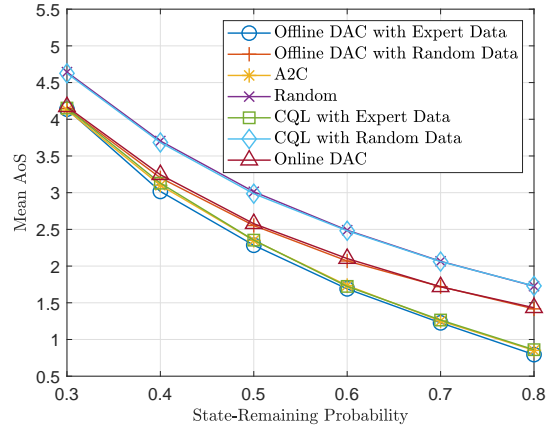}
    \caption{Mean AoS performance for the SN versus state-remaining probability: $I = 75$ and $\varphi = 0.5$.}
    \label{simu02_01}
\end{figure}

\begin{figure}[t]
    \centering
    \includegraphics[width=19pc]{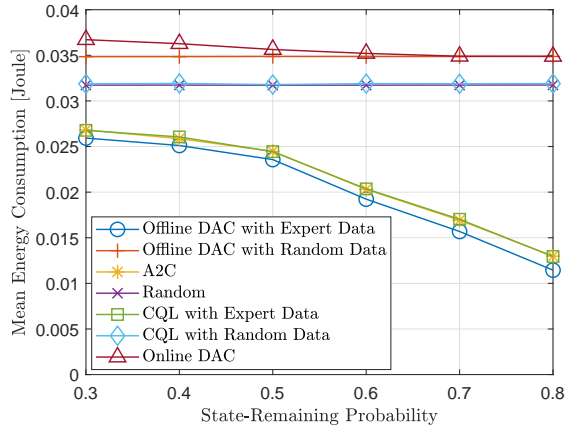}
    \caption{Mean energy consumption for the SN versus state-remaining probability: $I = 75$ and $\varphi = 0.5$.}
    \label{simu02_02}
\end{figure}

\begin{figure}[t]
    \centering
    \includegraphics[width=19pc]{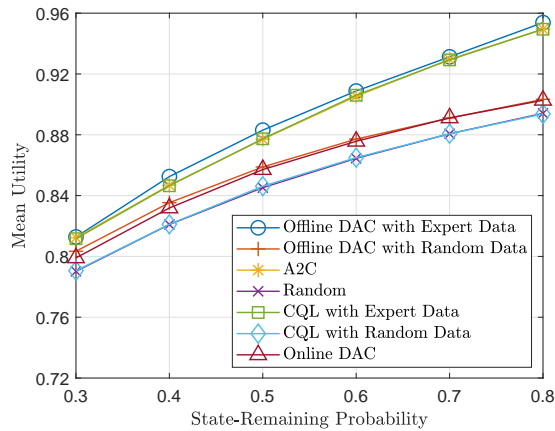}
    \caption{Mean utility performance for the SN versus state-remaining probability: $I = 75$ and $\varphi = 0.5$.}
    \label{simu02_03}
\end{figure}

By comparison with the baselines, we then move to demonstrate the performance of our proposed offline DAC scheme in terms of mean AoS, mean energy consumption and mean utility for the SN.
The proposed offline DAC and the CQL schemes are trained using both Expert Data and Random Data.
We configure a communication system similar as in the previous experiment except that for the process of interest at the SN, the state-remaining probability $\chi$ varies between $0.3$ and $0.8$.
The experimental results are exhibited in Figs. \ref{simu02_01}, \ref{simu02_02} and \ref{simu02_03}, which illustrate, respectively, the mean AoS, the mean energy consumption and the mean utility from all the online and offline schemes.

When being trained with Expert Data, it can be observed from the curves in Fig. \ref{simu02_03} that the proposed offline DAC scheme achieves the best mean utility performance, while the CQL scheme has nearly the same mean utility performance as the A2C scheme.
The increase in the state-remaining probability $\chi$ increases the chance for the destination to maintain a perfect inference of the process status.
As such, the mean AoS performance decreases, and the mean energy consumption of the SN also decreases due to the reduced process sampling frequency, as illustrated in Figs. \ref{simu02_01} and \ref{simu02_02}.
The mean energy consumption from the Random scheme keeps unchanged, which can be explained by the fully random process status sampling and RS selection.
When being trained with Random Data, Fig. \ref{simu02_03} reveals that our proposed offline DAC scheme slightly outperforms the online DAC scheme in terms of mean utility.
Different from the Random scheme, the proposed offline DAC scheme increases the process sampling frequency in order to bring down the mean AoS, but still converges to a random policy, which can be obviously seen from Figs. \ref{simu02_01} and \ref{simu02_02}.
That is, the proposed offline DAC scheme fails to dig out the optimal control policy from Random Data.
However, the CQL scheme performs worst as the Random scheme.
This corroborates that the CQL scheme merely imitates the control policies generating Expert Data and Random Data.

\subsubsection{Robustness to Dataset Quality}

\begin{figure}[t]
    \centering
    \includegraphics[width=19pc]{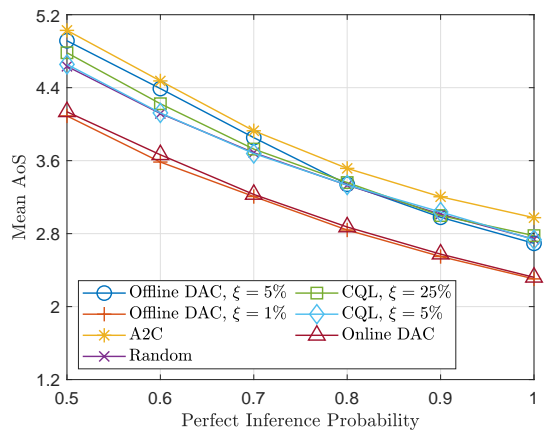}
    \caption{Mean AoS performance for the SN versus perfect inference probability: $I = 25$ and $\chi = 0.3$.}
    \label{simu03_01}
\end{figure}

\begin{figure}[t]
    \centering
    \includegraphics[width=19pc]{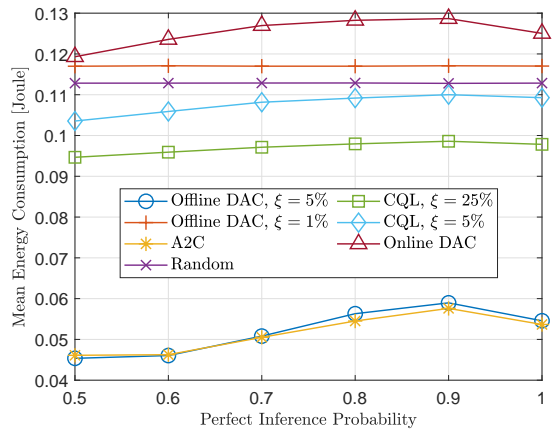}
    \caption{Mean energy consumption for the SN versus perfect inference probability: $I = 25$ and $\chi = 0.3$.}
    \label{simu03_02}
\end{figure}

\begin{figure}[t]
    \centering
    \includegraphics[width=19pc]{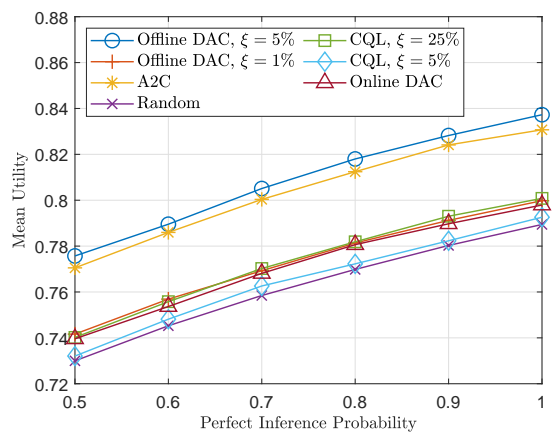}
    \caption{Mean utility performance for the SN versus perfect inference probability: $I = 25$ and $\chi = 0.3$.}
    \label{simu03_03}
\end{figure}

Finally, we carry out an experiment to examine the robustness of our proposed offline DAC scheme to the dataset quality.
In this experiment, we mix Expert Data with Random Data and let $\xi$ denote the fraction of Expert Data in a dataset used for training offline schemes.
We fix the number of reflecting elements and the state-remaining probability to $I = 25$ and $\chi = 0.3$, respectively.
Figs. \ref{simu03_01}, \ref{simu03_02} and \ref{simu03_03} depict the mean AoS, the mean energy consumption and the mean utility performance of the SN from all schemes.

It is apparent from Fig. \ref{simu03_01} that as the perfect inference probability increases, the mean AoS decreases.
The larger the probability of the perfect inference on a reconstructed status update, the destination is updated with more fresh semantics of the process of interest, inspiring the SN to sample more frequently.
This explains why the mean energy consumptions from the offline DAC scheme trained using datasets with $\xi = 5\%$ Expert Data, the CQL scheme and the online DAC scheme increase, as perceived in Fig. \ref{simu03_02}.
When the perfect inference probability is sufficiently large, there is no need for the SN to retain a high process sampling frequency, resulting in the reduction of mean energy consumption.
When the fraction of Expert Data is $\xi = 1\%$, it is challenging to distinguish the dataset from Random Data.
The proposed offline DAC scheme converges to a random RS selection policy, with which the SN consumes the mean energy at a steady level.
By comparing Figs. \ref{simu02_02} and \ref{simu03_02}, we also discover that the SN with the Random scheme consumes less energy when the IRS is equipped with a larger number of reflecting elements, justifying the energy efficiency improvements by the IRS.
Given the weighting factors, the AoS dominates the utility function value, which conforms the mean utility performance trends in Fig. \ref{simu03_03}.
Last but not least, the mean utility performance from the proposed offline DAC scheme trained using datasets with only $\xi = 5\%$ Expert Data moderately outperforms the A2C scheme and significantly outperforms the CQL baseline trained using datasets with $\xi = 25\%$ Expert Data.
From this experiment, the proposed offline DAC scheme exhibits the highly strong robustness to the dataset quality.
On the contrary, the CQL scheme is sensitive to the quality of a dataset, which is in line with the findings from \cite{Kumar20}.

\section{Conclusions}
\label{conc}

In this paper, we propose the notion of AoS to investigate the semantics freshness under the context of an IRS-assisted cooperative relay communication system.
Considering the system uncertainties, we formulate the problem of joint process status sampling and RS selection as an MDP, where the objective of the SN is to maximize the expected discounted utility performance over the discrete time slots.
We first develop an online on-policy DAC scheme to alleviate the dependence on the MDP statistics.
To address the ``chicken and egg'' paradox faced by the online DAC scheme, we then derive an offline DAC scheme.
The proposed offline DAC scheme efficiently lower-bounds the estimated Q-function of OOD actions, without any further interactions with the communication system.
The accuracy of the proposed studies is theoretically verified.
Furthermore, the numerical experiments confirm that the proposed offline DAC outperforms the state-of-the-art baselines in terms of mean utility and is highly robust to dataset quality.


\begin{thebibliography}{29}

\bibitem{Wang13}
R. Wang and V. K. N. Lau, ``Delay-aware two-hop cooperative relay communications via approximate MDP and stochastic learning,'' \emph{IEEE Trans. Inf. Theory}, vol. 59, no. 11, pp. 7645--7670, Nov. 2013.

\bibitem{ZKang22}
Z. Kang, C. You, and R. Zhang, ``IRS-aided wireless relaying: Deployment strategy and capacity scaling,'' \emph{IEEE Wireless Commun. Lett.}, vol. 11, no. 2, pp. 215--219, Feb. 2022.

\bibitem{Bjor20}
E. Bj\"{o}rnson, \"{O}. \"{O}zdogan, and E. G. Larsson, ``Intelligent reflecting surface versus decode-and-forward: How large surfaces are needed to beat relaying?,'' \emph{IEEE Wireless Commun. Lett.}, vol. 9, no. 2, pp. 244--248, Feb. 2020.

\bibitem{Noor22}
M. Noor-A-Rahim, F. Firyaguna, J. John, M. O. Khyam, D. Pesch, E. Armstrong, H. Claussen, and H. V. Poor, ``Towards Industry 5.0: Intelligent reflecting surface (IRS) in smart manufacturing,'' \emph{IEEE Commun. Mag.}, Early Access Article, 2022.

\bibitem{Wang21}
Y. Wang, W. Wang, D. Liu, X. Jin, J. Jiang, and K. Chen, ``Enabling edge-cloud video analytics for robotics applications,'' in \emph{Proc. IEEE INFOCOM}, Vancouver, BC, Canada, May 2021.

\bibitem{Du20}
K. Du, A. Pervaiz, X. Yuan, A. Chowdhery, Q. Zhang, H. Hoffmann, and J. Jiang, ``Server-driven video streaming for deep learning inference,'' in \emph{Proc. ACM SIGCOMM}, Online, Aug. 2020.

\bibitem{Kang22}
X. Kang, B. Song, J. Guo, Z. Qin, and F. R. Yu, ``Task-oriented image transmission for scene classification in unmanned aerial systems,'' \emph{IEEE Trans. Commun.}, vol. 70, no. 8, pp. 5181--5192, Aug. 2022.

\bibitem{Kaul12}
S. Kaul, R. Yates, and M. Gruteser, ``Real-time status: How often should one update?,'' in \emph{Proc. IEEE INFOCOM}, Orlando, FL, USA, Mar. 2012.

\bibitem{ChenX20}
X. Chen, C. Wu, T. Chen, H. Zhang, Z. Liu, Y. Zhang, and M. Bennis, ``Age of information aware radio resource management in vehicular networks: A proactive deep reinforcement learning perspective,'' \emph{IEEE Trans. Wireless Commun.}, vol. 19, no. 4, pp. 2268--2281, Apr. 2020.

\bibitem{ChenX22}
X. Chen, C. Wu, T. Chen, Z. Liu, H. Zhang, M. Bennis, H. Liu, and Y. Ji, ``Information freshness-aware task offloading in air-ground integrated edge computing systems,'' \emph{IEEE J. Sel. Areas Commun.}, vol. 40, no. 1, pp. 243--258, Jan. 2022.

\bibitem{Qian20}
Z. Qian, F. Wu, J. Pan, K. Srinivasan, and N. B. Shroff, ``Minimizing age of information in multi-channel time-sensitive information update systems,'' in \emph{Proc. IEEE INFOCOM}, Toronto, ON, Canada, Jul. 2020.

\bibitem{Abd20}
M. A. Abd-Elmagid, H. S. Dhillon, and N. Pappas, ``A reinforcement learning framework for optimizing age of information in RF-powered communication systems,'' \emph{IEEE Trans. Commun.}, vol. 68, no. 8, pp. 4747--4760, Aug. 2020.

\bibitem{Ahani22}
G. Ahani, D. Yuan, and S. Sun, ``Optimal scheduling of age-centric caching: Tractability and computation,'' \emph{IEEE Trans. Mobile Comput.}, vol. 21, no. 8, pp. 2939--2954, 1 Aug. 2022.

\bibitem{Yates21}
R. D. Yates, Y. Sun, D. R. Brown, S. K. Kaul, E. Modiano, and S. Ulukus, ``Age of information: An introduction and survey,'' \emph{IEEE J. Sel. Areas Commun.}, vol. 39, no. 5, pp. 1183--1210, May 2021.

\bibitem{Talak17}
R. Talak, S. Karaman, and E. Modiano, ``Minimizing age-of-information in multi-hop wireless networks,'' in \emph{Proc. Allerton}, Monticello, IL, USA, Oct. 2017,

\bibitem{Farazi18}
S. Farazi, A. G. Klein, J. A. McNeill, and D. Richard Brown, ``On the age of information in multi-source multi-hop wireless status update networks,'' in \emph{Proc. IEEE SPAWC}, Kalamata, Greece, Jun. 2018.

\bibitem{He22}
T. He, K.-W. Chin, Z. Zhang, T. Liu, and J. Wen, ``Optimizing information freshness in RF-powered multi-hop wireless networks,'' \emph{IEEE Trans. Wireless Commun.}, Eerly Access Article, 2022.

\bibitem{Lou22}
J. Lou, X. Yuan, P. Sigdel, X. Qin, S. Kompella, and N.-F. Tzeng, ``Age of information optimization in multi-channel based multi-hop wireless networks,'' \emph{IEEE Trans. Mobile Comput.}, Early Access Article, 2022.

\bibitem{Liu22}
Q. Liu, H. Zeng, and M. Chen, ``Minimizing AoI with throughput requirements in multi-path network communication,'' \emph{IEEE/ACM Trans. Netw.}, vol. 30, no. 3, pp. 1203--1216, Jun. 2022.

\bibitem{Chen19J}
X. Chen, H. Zhang, C. Wu, S. Mao, Y. Ji, and M. Bennis, ``Optimized computation offloading performance in virtual edge computing systems via deep reinforcement learning,'' \emph{ IEEE Internet Things J.}, vol. 6, no. 3, pp. 4005--4018, Jun. 2019.

\bibitem{Rich98}
R. S. Sutton and A. G. Barto, \emph{Reinforcement Learning: An Introduction}. Cambridge, MA: MIT Press, 1998.

\bibitem{Gu21}
Y. Gu, Q. Wang, H. Chen, Y. Li, and B. Vucetic, ``Optimizing information freshness in two-hop status update systems under a resource constraint," \emph{IEEE J. Sel. Areas Commun.}, vol. 39, no. 5, pp. 1380--1392, May 2021.

\bibitem{Tripathi21}
V. Tripathi, R. Talak, and E. Modiano, ``Information freshness in multi-hop wireless networks,'' \emph{arXiv}, Nov. 2021. [Online]. Available: https://arxiv.org/pdf/2111.09217.pdf [Accessed: 13 Aug. 2022].

\bibitem{Sun20}
Y. Sun, Y. Polyanskiy, and E. Uysal, ``Sampling of the wiener process for remote estimation over a channel with random delay,'' \emph{IEEE Trans. Inf. Theory}, vol. 66, no. 2, pp. 1118--1135, Feb. 2020.

\bibitem{Maat20}
A. Maatouk, S. Kriouile, M. Assaad, and A. Ephremides, ``The age of incorrect information: A new performance metric for status updates,'' \emph{IEEE/ACM Trans. Netw.}, vol. 28, no. 5, pp. 2215--2228, Oct. 2020.

\bibitem{Xu21}
Z. Xu, K. Wu, W. Zhang, J. Tang, Y. Wang, and G. Xue, ``PnP-DRL: A plug-and-play deep reinforcement learning approach for experience-driven networking,'' \emph{IEEE J. Sel. Areas Commun.}, vol. 39, no. 8, pp. 2476--2486, Aug. 2021.

\bibitem{Levine20}
S. Levine, A. Kumar, G. Tucker, and J. Fu, ``Offline reinforcement learning: Tutorial, review, and perspectives on open problems,'' \emph{arXiv}, Nov. 2020. [Online]. Available: https://arxiv.org/pdf/2005.01643.pdf [Accessed: 16 Aug. 2022].

\bibitem{Fujimoto19}
S. Fujimoto, D. Meger, and D. Precup, ``Off-policy deep reinforcement learning without exploration,'' in \emph{Proc. ICML}, Long Beach, CA, USA, Jun. 2019.

\bibitem{Kumar19}
A. Kumar, J. Fu, G. Tucker, and S. Levine, ``Stabilizing off-policy Q-learning via bootstrapping error reduction,'' in \emph{Proc. NIPS}, Vancouver, Canada, Dec. 2019.

\bibitem{Siegel20}
N. Siegel, J. T. Springenberg, F. Berkenkamp, A. Abdolmaleki, M. Neunert, T. Lampe, R. Hafner, N. Heess, and M. Riedmiller, ``Keep doing what worked: Behavior modelling priors for offline reinforcement learning,'' in \emph{Proc. ICLR}, Virtual, Apr. 2020.

\bibitem{Wang20}
Z. Wang, A. Novikov, K. \.{Z}o{\l}na, J. T. Springenberg, S. Reed, B. Shahriari, N. Siegel, J. Merel, C. Gulcehre, N. Heess, and N. de Freitas, ``Critic regularized regression,'' in \emph{Proc. NIPS}, Virtual, Dec. 2020.

\bibitem{Kumar20}
A. Kumar, A. Zhou, G. Tucker, and S. Levine, ``Conservative Q-learning for offline reinforcement learning,'' in \emph{Proc. NIPS}, Virtual, Dec. 2020.

\bibitem{Xu22}
H. Xu, X. Zhan, and X. Zhu, ``Constraints penalized Q-learning for safe offline reinforcement learning,'' in \emph{Proc. AAAI}, Virtual, Feb.-Mar. 2022.

\bibitem{Mnih16}
V. Mnih, A. P. Badia, M. Mirza, A. Graves, T. Lillicrap, T. Harley, D. Silver, and K. Kavukcuoglu, ``Asynchronous methods for deep reinforcement learning,'' in \emph{Proc. ICML}, New York, NY, USA, Jul. 2016.

\bibitem{Cho09}
S. Cho, E. W. Jang, and J. M. Cioffi, ``Handover in multihop cellular networks,'' \emph{IEEE Commun. Mag.}, vol. 47, no. 7, pp. 64--73, Jul. 2009.

\bibitem{Zlat19}
N. Zlatanov and R. Schober, ``Buffer-aided relaying with adaptive link selection-fixed and mixed rate transmission,'' \emph{IEEE Trans. Inf. Theory}, vol. 59, no. 5, pp. 2816--2840, May 2013.

\bibitem{Wang22}
S. Wang, M. Chen, Z. Yang, C. Yin, W. Saad, S. Cui, and H. V. Poor, ``Distributed reinforcement learning for age of information minimization in real-time IoT systems,'' \emph{IEEE J. Sel. Top. Signal Process.}, vol. 16, no. 3, pp. 501--515, Apr. 2022.

\bibitem{Sutt99}
R. S. Sutton, D. Precup, and S. Singh, ``Between MDPs and semi-MDPs: A framework for temporal abstraction in reinforcement learning,'' \emph{Artif. Intell.}, vol. 112, no. 1--2, pp. 181--211, Aug. 1999.

\bibitem{Fiedler10}
M. Fiedler, T. Hossfeld, and P. Tran-Gia, ``A generic quantitative relationship between quality of experience and quality of service,'' \emph{IEEE Netw.}, vol. 24, no. 2, pp. 36--41, Mar./Apr. 2010.

\bibitem{Maha96}
S. Mahadevan, ``Sensitive discount optimality: Unifying discounted and average reward reinforcement learning,'' in \emph{Proc. ICML}, Bari, Italy, Jul. 1996.

\bibitem{Tsit07}
J. N. Tsitsiklis, ``NP-hardness of checking the unichain condition in average cost MDPs,'' \emph{Oper. Res. Lett.}, vol. 35, no. 3, pp. 319--323, May 2007.

\bibitem{Eyes22}
B. Eysenbach and S. Levine, ``Maximum entropy RL (provably) solves some robust RL problems,'' in \emph{Proc. ICLR}, Virtual, Apr. 2022.

\bibitem{Haar18}
T. Haarnoja, A. Zhou, P. Abbeel, and S. Levine, ``Soft actor-critic: Off-policy maximum entropy deep reinforcement learning with a stochastic actor,'' in \emph{Proc. ICML}, Stockholm, Sweden, Jul. 2018.

\bibitem{Bellman57}
R. Bellman, \emph{Dynamic Programming}. Princeton, NJ: Princeton University Press, 1957.

\bibitem{Mnih15}
V. Mnih, K. Kavukcuoglu, D. Silver, A. A. Rusu, J. Veness, M. G. Bellemare, A. Graves, M. Riedmiller, A. K. Fidjeland, G. Ostrovski, S. Petersen, C. Beattie, A. Sadik, I. Antonoglou, H. King, D. Kumaran, D. Wierstra, S. Legg, and D. Hassabis, ``Human-level control through deep reinforcement learning,'' \emph{Nature}, vol. 518, no. 7540, pp. 529--533, Feb. 2015.

\bibitem{Chen19O}
X. Chen, Z. Zhao, C. Wu, M. Bennis, H. Liu, Y. Ji, and H. Zhang, ``Multi-tenant cross-slice resource orchestration: A deep reinforcement learning approach,'' \emph{IEEE J. Sel. Areas Commun.}, vol. 37, no. 10, pp. 2377--2392, Oct. 2019.

\bibitem{Dula21}
G. Dulac-Arnold, N. Levine, D. J. Mankowitz, J. Li, C. Paduraru, S. Gowal, and T. Hester, ``Challenges of real-world reinforcement learning: Definitions, benchmarks and analysis,'' \emph{Mach. Learn.}, vol. 110, pp. 2419--2468, Apr. 2021.

\bibitem{Wu21}
Y. Wu, K. Zhang, and Y. Zhang, ``Digital twin networks: A survey,'' \emph{ IEEE Internet Things J.}, vol. 8, no. 18, pp. 13789--13804, Sep. 2021.

\bibitem{Abdolmaleki18}
A. Abdolmaleki, J. T. Springenberg, N. Heess, Y. Tassa, and R. Munos, ``Maximum a posteriori policy optimisation,'' in \emph{Proc. ICLR}, Vancouver, BC, Canada, Apr.-May 2018.

\bibitem{Su20}
D. Su, J. Ooi, T. Lu, D. Schuurmans, and C. Boutilier, ``ConQUR: Mitigating delusional bias in deep Q-learning,'' in \emph{Proc. ICML}, Virtual, Jul. 2020.

\bibitem{Lin20}
K. Lin and J. Zhou, ``Ranking policy gradient,'' in \emph{Proc. ICLR}, Virtual, Apr. 2020.

\bibitem{Rezaeifar22}
S. Rezaeifar, R. Dadashi, N. Vieillard, L. Hussenot, O. Bachem, O. Pietquin, and M. Geist, ``Offline reinforcement learning as anti-exploration,'' in \emph{Proc. AAAI}, Virtual, Feb.-Mar. 2022.

\bibitem{Donoghue21}
B. O'Donoghue, ``Variational bayesian reinforcement learning with regret bounds,'' in \emph{Proc. NeurIPS}, Virtual, Dec. 2021.

\bibitem{Min22}
Y. Min, J. He, T. Wang, and Q. Gu, ``Learning stochastic shortest path with linear function approximation,'' in \emph{Proc. ICML}, Baltimore, MD, USA, Jul. 2022.

\bibitem{Sharma22}
A. Sharma, R. Ahmad, and C. Finn,``A state-distribution matching approach to non-Episodic reinforcement learning,'' in \emph{Proc. ICML}, Baltimore, MD, USA, Jul. 2022.

\bibitem{Nachum17}
O. Nachum, M. Norouzi, K. Xu, and D. Schuurmans, ``Bridging the gap between value and policy based reinforcement learning,'' in \emph{Proc. NIPS}, Long Beach, CA, USA, Dec. 2017.

\bibitem{Haarnoja17}
T. Haarnoja, H. Tang, P. Abbeel, and S. Levine, ``Reinforcement learning with deep energy-based policies,'' in \emph{Proc. ICML}, Sydney, Australia, Aug. 2017.

\bibitem{Nair10}
V. Nair and G. E. Hinton, ``Rectified linear units improve restricted boltzmann machines,'' in \emph{Proc. ICML}, Haifa, Israel, Jun. 2010.

\bibitem{Goodfellow16}
I. Goodfellow, Y. Bengio, and A. Courville, \emph{Deep Learning}. Cambridge, MA: MIT Press, 2016.

\bibitem{King15}
D. P. Kingma and J. Ba, ``Adam: A Method for Stochastic Optimization,'' in \emph{Proc. ICLR}, San Diego, CA, May 2015.


\end{thebibliography}
\end{document}